\documentclass[conference]{IEEEtran}
\usepackage{booktabs} % For formal tables

\usepackage[linesnumbered,ruled,vlined]{algorithm2e}
\usepackage{algpseudocode}
\usepackage[colorlinks,linkcolor=blue]{hyperref}
\usepackage{subfigure}
\usepackage{graphicx}
\usepackage{amsmath}
\usepackage{bm}
\usepackage{url}
\usepackage{stmaryrd}
\usepackage{balance}
\usepackage{amssymb}
\usepackage{pgfplots}
\usepackage{pifont}
\usepackage{epsfig}
\usepackage{booktabs}% professional tables
\usepackage{pgffor}
\usepackage{tikz}
\usepackage{multirow}
\usepackage{amsmath}

\usepackage{tikz}
\usetikzlibrary{positioning,arrows,calc}
\usepackage[all]{xy}
\usepackage{makecell}
\usepackage{enumitem}
\newcommand{\our}{\textsc{EnsemFDet}}

\newcommand{\ouralgorithm}{\textsc{FDet}}
\newcommand{\ourfixedk}{\textsc{EnsemFDet-Fix-K}}
\newcommand{\spoken}{\textsc{SPOKEN}}
\newcommand{\Fbox}{\textsc{FBox}}
\newcommand{\Fraudar}{\textsc{Fraudar}}

\newcommand{\RNum}[1]{\uppercase\expandafter{\romannumeral #1\relax}}
\begin{document}
	
	%%
	%% The "title" command has an optional parameter,
	%% allowing the author to define a "short title" to be used in page headers.
	\title{EnsemFDet: An Ensemble Approach to Fraud Detection based on Bipartite Graph}
	
	%%
	%% The "author" command and its associated commands are used to define
	%% the authors and their affiliations.
	%% Of note is the shared affiliation of the first two authors, and the
	%% "authornote" and "authornotemark" commands
	%% used to denote shared contribution to the research.
	
	%\author{\IEEEauthorblockN{Yuxiang Ren\footnote{This work was done while the author was an intern at JD Digits.}}
	%	\IEEEauthorblockA{\textit{IFM Lab} \\
	%		\textit{Florida State University}\\
	%		Tallahassee, USA \\
	%		yuxiang@ifmlab.org}
	%	\and
	%	\IEEEauthorblockN{Hao Zhu}
	%	\IEEEauthorblockA{\textit{JD Digits} \\
	%		Beijing, China \\
	%		zhuhao9@jd.com}
	%	\and
	%	\IEEEauthorblockN{Jiawei Zhang}
	%	\IEEEauthorblockA{\textit{IFM Lab} \\
	%		\textit{Florida State University}\\
	%		Tallahassee, USA \\
	%		jiawei@ifmlab.org}
	%	\and
	%	\IEEEauthorblockN{Peng Dai}
	%	\IEEEauthorblockA{\textit{AI Lab, JD Digits} \\
	%		Mountain View, USA \\
	%		peng.dai@jd.com}
	%	\and
	%	\IEEEauthorblockN{Liefeng Bo}
	%	\IEEEauthorblockA{\textit{AI Lab, JD Digits} \\
	%		Mountain View, USA \\
	%		liefeng.bo@jd.com}
	%	
	%}
	\author{\IEEEauthorblockN{Yuxiang Ren\IEEEauthorrefmark{1},
			Hao Zhu\IEEEauthorrefmark{2}, Jiawei Zhang\IEEEauthorrefmark{1}
			, Peng Dai\IEEEauthorrefmark{3} and
			Liefeng Bo\IEEEauthorrefmark{3}}
		\IEEEauthorblockA{\IEEEauthorrefmark{1}IFM Lab, Department of Computer Science, Florida State University, FL, USA\\
			\IEEEauthorrefmark{2}College of
			Engineering \& Computer Science, Australian National University,  Canberra, Australia\\
			\IEEEauthorrefmark{3}AI Lab, JD Finance America Corporation, CA, USA\\
			Email: yuxiang@ifmlab.org,
			hao.zhu@anu.edu.au,
			jiawei@ifmlab.org,
			peng.dai@jd.com,
			liefeng.bo@jd.com}}
	
	\maketitle

	\begin{abstract}
		Fraud detection is extremely critical for e-commerce business platforms. It is the intent of the companies to detect and prevent fraud as early as possible. Utilizing graph structure data and identifying unexpected dense subgraphs as suspicious is a category of commonly used fraud detection methods. Among them, spectral methods solve the problem efficiently but hurt the performance due to the relaxed constraints. Besides, existing heuristic methods cannot be accelerated with parallel computation and fail to control the scope of returned suspicious nodes because they provide a set of subgraphs with diverse node sizes. These drawbacks affect the real-world applications of existing methods. In this paper, we propose an \textit{\textbf{Ensem}ble based \textbf{F}raud \textbf{DET}ection} ({\our}) method to scale up promotional campaigns fraud detection in bipartite graphs by decomposing the original problem into subproblems on small-sized subgraphs. By oversampling the graph and solving the subproblems, the ensemble approach further votes suspicious nodes without sacrificing the prediction accuracy. Extensive experiments have been done on real transaction data from \textbf{JD.com}, which is one of the world's largest e-commerce platforms. Experimental results demonstrate the effectiveness, practicability, and scalability of {\our}. More specifically, {\our} is up to 100x faster than the state-of-the-art methods due to its parallelism with all aspects of data.
	\end{abstract}
\maketitle
\section{Introduction}\label{sec:introduction}

As online services are becoming increasingly popular, it attracts fraudsters to look for measures to abuse their virtual currency system~\cite{cao2017hitfraud,oentaryo2014detecting}. For example, a large number of accounts are created and controlled by a group of fraudsters, and these malicious fraudsters may cash out and obtain ill-gotten wealth, thereby greatly damaging the entire virtual currency system. 

Among many fraudulent activities, arbitrage frauds that use online promotional campaigns are common but not easy to deal with. E-commerce platforms usually launch many promotional campaigns to increase platform clickstream and attract user consumption.
For example, each transaction can enjoy a 5-dollar discount if the transaction cost is greater than \$5.01, or the discount can be a random value within 5 dollars no matter what the total transaction cost is. Fraudsters usually register many malicious accounts in batches and control malicious accounts to make bulk purchases during the promotion period to illegally profit. This kind of fraud not only damages the company's interests, where the costs invested in promotional activities are not converted into clickstreams, but also deprives many users of the discounts they should have received. However, fraud behaviors in promotional campaigns are not easy to detect and defend. Most promotional campaigns will not last long and change a lot next time. According to our observation of the promotional campaigns on JD.com, most fraudulent accounts used by fraudsters will not be reused after a period of time, and the features of fraud behaviors will also change with the different promotional campaigns. Therefore, it is difficult to label fraud behaviors and utilize historical data. In other words, feature-based supervised learning \cite{wang2019fdgars, ZCXJXL18, gomez2018end, park2019fraud}, a classical way to detect fraud behaviors, is very hard to be applied in such a scenario. In order to reduce the growing abuse of promotional campaigns, expert-rule systems are developed to identify suspicious accounts. However, fraudsters would update means of fraudulence using techniques such as device and IP obfuscation to evade the rules, and there are increasing difficulties in detecting these more sophisticated attacks. In addition, millions of fake user accounts are being created every day, and they further reduce the density of attacking and thus evade related rules. Moreover, expert-rule systems heavily depend on the expert experiences finding in manual monitoring.

Compared with some camouflage measures to bypass expert-rule systems, the traces (e.g., links and nodes) left by fraudsters on graphs are difficult to eliminate. Graph-based methods \cite{prakash2010eigenspokes,shah2014spotting,jiang2014inferring,hooi2016fraudar} can comprehensively consider all events and users. In this way, they can detect fraud behaviors and promotion abuse systematically. By finding the hidden relationships and behavior consistency among all accounts, graph-based methods can detect different groups of fraudsters without any label for training which is extremely expensive and time-consuming to get in the real world. As data-driven approaches, graph-based methods can reduce the costs generated by the lagged effect of making rules as well.

Graph-based methods detect groups of fraudsters in promotional campaigns by identifying unexpectedly dense regions of the graph. We can represent the transaction information of promotional campaigns as a bipartite graph, which can be defined as the 'who buy-from where' graph. The nodes on two sides of the 'who buy-from where' graph are users and merchants respectively. The connections between users and merchants denote the purchase behavior. Based on the 'who buy-from where' graph, there exist two clues about the fraudsters according to our observations of historical transaction data during promotional campaigns on JD.com:
\begin{itemize}[leftmargin=*]
	\item \textit{Synchronized Behavior}: Fraudsters are limited by the time of promotional campaigns. Basically, Fraudsters need to achieve a specific goal in a short term. Therefore, suspicious nodes have extremely synchronized behavior patterns within a short time, because they are often required to perform similar tasks together such as payments in specific stores.
	\item \textit{Rare Behavior}: The connectivity patterns among high suspicious nodes are very different from the majority. Usually, the density of subgraphs composed of highly suspicious nodes is significantly higher than other parts of the full graph.
\end{itemize}

Intuitively, graph-based fraud detection methods share the same idea with density-based clustering approaches: both looking for subgraphs with a higher density than the remainder of the graphs/data. The task of graph-based fraud detection is to find all subgraphs that have anomalous patterns (usually of high density) from the provided objective graph. There are two main types of approaches for this task: spectral methods \cite{prakash2010eigenspokes,shah2014spotting,jiang2014inferring} and heuristic algorithms \cite{hooi2016fraudar,tsourakakis2013denser}. Spectral methods solve the problem efficiently but hurt the performance due to the complicated relaxed constraints design~\cite{wang2010flexible}. Compared with spectral methods, the heuristic algorithms commonly have better performance. However, heuristic methods commonly detect suspicious subgraphs with diverse sizes, and all nodes in one suspicious subgraph will be labeled as suspicious nodes. This property hurts the practicability of heuristic methods in real-world applications because they will lead to the zigzag ROC curve and the zigzag precision-recall curve. Therefore, we cannot select suspicious nodes by configuring the recall or false positive rate. Besides, heuristic methods commonly tend to return the densest block because there is no efficient strategy to control how many blocks the algorithms should find. Through our analysis of historical transaction data on JD.com, it is found that there are usually multiple groups of fraudsters in the same period of promotional campaigns. The fraudsters of different groups are reflected in the 'who buy-from where' graph, that is, there are high-density disjoint subgraphs. Instead of identifying the densest subgraph, the target of the fraud detection problem is not only finding the densest subgraph but also extracting all other unexpected dense subgraphs. The number of dense subgraphs is normally a hyper-parameter for heuristic methods. The most important thing is that the heuristic process in algorithms is sequential and thus it is difficult to be parallelized. These drawbacks greatly hinder their applications in the real world. 

To address the aforementioned drawbacks, we propose a model, namely  \textit{\textbf{Ensem}ble based \textbf{F}raud \textbf{DET}ection} ({\our}), in this paper. 
%In {\our}, we first define the optimization problem of graph-based fraud detection corresponding to our business scenarios. Second, we show how to implement graph-based fraud detection practical by decomposing the original graph into small-sized subgraphs so as to lower the search costs. 
An ensemble framework is utilized in {\our}, which employs three structural sampling methods for the bipartite graph to decompose the original graph into small-sized graphs so as to lower the search costs. The ensemble framework aggregating the outputs of several subproblems can reduce the overall risk of achieving a particularly poor suboptimal solution, and thus maintaining high prediction accuracy. We propose a fraud detection method {\ouralgorithm} working on the sampled small-sized graphs to detect fraudsters. {\ouralgorithm} enables a more efficient search for the number of fraud subgraphs existing on the original graph, where the hyper-parameter (i.e., the number of fraud subgraphs) can be determined by a novel strategy automatically. In addition, {\our} can compute fraud detection in sampled graphs in parallel, thus accelerating the detection process. 
%Using real-life datasets of promotion campaigns, we conduct extensive experiments and show that {\our} for fraud detection is effective, practical, scalable and stable. 

The contributions of our work are summarized as follows:
\begin{itemize}[leftmargin=*]
	\item We define the optimization problem of graph-based fraud detection corresponding to our promotional campaign scenarios.
	\item We propose a novel ensemble-based framework {\our} which implements promotional campaign fraud detection by decomposing the original graph into small-sized graphs so as to lower the search costs. As the core of {\our}, the algorithm {\ouralgorithm} is proposed to detect fraudsters on small-sized graphs and determine the number of fraud subgraphs automatically.   
	\item We conduct extensive experiments on three real-world datasets based on different promotional campaigns. The experimental results show that {\our} for fraud detection is effective, practical, scalable and stable. 
\end{itemize}
The remaining paper is organized as follows. We review the related works in Section~\ref{sec:related_work}. Then we introduce several important concepts and formulate the problem in Section~\ref{sec:formulation}. The proposed model {\our} is introduced in Section~\ref{sec:method}, whose effectiveness is evaluated in Section~\ref{sec:experiment}. Finally, we conclude this paper in Section~\ref{sec:conclusion}.

%-----------------------------------------------
\section{Related Work} \label{sec:related_work}

Approaches in the field of fraud detection can be classified into two categories: feature-based and graph-based.
\subsection{Feature-based Method}
Feature-based methods model the users and activities by representing them through a set of attributes, such that each entity is represented as a vector in a multi-dimensional space. In order to distinguish normal users from fraudsters, the appropriate set of attributes should be designed hand-crafted, so that entities will lie in significantly different regions in this feature space. With the rise of deep learning, neural networks have also been used to extract features from fraudsters and relating events, and accomplish fraud detection through a classifier~\cite{wang2019fdgars,ZCXJXL18,li2019spam,gomez2018end,park2019fraud,rayana2015collective}. This class of algorithms often require pre-labeled training data, which are used to find the distinguishing attributes. These algorithms have been
successfully used to predict trolls~\cite{cheng2015antisocial}, vandals~\cite{kumar2015vews}, hoaxes~\cite{kumar2016disinformation}, and sockpuppets~\cite{kumar2017army}, among several other anomaly entities~\cite{kaghazgaran2019wide}.
However, obtaining reliable labels is extremely expensive and time-consuming in real business scenarios, which may limit us from detecting the fraud in the very early stage. Some of these feature-based algorithms are also efficient and effective to find anomaly users from unlabeled training data like linear models \cite{ma2003time}, proximity-based models \cite{breunig2000lof}, probabilistic models \cite{kriegel2008angle} and ensemble models \cite{liu2008isolation}. 

\subsection{Graph-based Method}
Graph-based methods are designed based on the intuition that anomalous entities act synchronously, i.e., they often take similar actions in near-similar times. When representing entities using their actions in a multi-dimensional space, the synchronous behavior of anomalous entities will form blocks with a higher density. Graph-based algorithms aim at identifying these dense-blocks in large-scale behavior logs. These algorithms have been successfully used to predict ill-gotten page Likes \cite{beutel2013copycatch}, zombie followers \cite{jiang2014catchsync}, money laundering~\cite{liflowscope}, and spam \cite{jiang2015general}. Graph-based approaches are more robust when facing adversarial attacking~\cite{hooi2016fraudar, dou2020robust}, because any strange behavior would unavoidably generate edges in the graph like 'who-buy-where' and 'who-buy-what'. There are two main types of approaches in graph-based algorithms: spectral methods and heuristic methods. Usually, graph-based methods identify unexpectedly dense regions of the graph and treat related nodes as suspicious. They try to transform the problem of identifying dense regions as the graph partitioning problem. Therefore, there are two different ways to solve the problem. First, by spectral relaxation, graph partitioning can be efficiently solved by SVD or eigenvalue decomposition. SPOKEN \cite{prakash2010eigenspokes} observes the 'eigenspokes' pattern produced by pairs of eigenvectors of graphs and is later generalized for fraud detection in \cite{jiang2014inferring}. FBOX \cite{shah2014spotting} focuses on the residual of SVD to detect attacks and \cite{jiang2015general} extends SVD to multimodal data. Several methods have used HITS-like \cite{kleinberg1999authoritative} ideas to detect fraud in graphs \cite{cao2012aiding,ghosh2012understanding,gyongyi2004combating,jiang2014catchsync,wu2006propagating}. Heuristic methods also solve the problem efficiently and effectively. Charikar et al. propose a heuristic method to find subgraphs with a large average degree \cite{charikar2000greedy}, which shows that subgraph average degree can be optimized with approximation guarantees. Tsourakakis et al.\cite{tsourakakis2013denser} extend \cite{charikar2000greedy} to the K-clique densest subgraph problem. Hooi et al.~\cite{hooi2016fraudar} propose Fraudar to deal with camouflages and provide upper bounds on the effectiveness of fraudsters. However, Fraudar detects fraud subgraphs of different sizes, and all nodes in a suspicious subgraph will be marked as fraud nodes. Moreover, Fraudar cannot automatically determine the number of suspicious subgraphs.
{\our} proposed in this paper can handle these drawbacks. At the same time, the parallelism of {\our} enables it to be more applicable to larger data.
% Compared with other approaches,

%\vspace{-8pt}
\section{Problem Formulation} \label{sec:formulation}
%\vspace{-4pt}

As an important part of marketing campaigns, e-payment usually provides many promotional campaigns with interesting services like a random discount. For example, each transaction can enjoy a \$5 discount if the transaction cost is greater than \$5.01, or the discount can be a random value within 5 dollars no matter what the total transaction cost is. We can represent the transaction information during promotional campaigns as a bipartite graph, which can be defined formally as:

\noindent \textbf{Definition 1}
('who buy-from where' graph): 
Consider user nodes $\mathcal{U}=\{u_1,...,u_m\}$, merchant nodes $\mathcal{V}=\{v_1,...,v_n\}$, and $\mathcal{E}$ which denotes the set of connections between $\mathcal{U}$ and $\mathcal{V}$ representing a purchase, we define that the bipartite graph $G = (\mathcal{U}\cup \mathcal{V},\mathcal{E})$ is a 'who buy-from where' graph.

%Based on the 'who buy-from where' graph, there exist two clues about the fraudsters according to our observations:
%\begin{itemize}[leftmargin=*]
%	\item \textit{synchronized behavior}: Fraudsters are limited by the time of promotional campaigns. Basically, Fraudsters need to achieve a specific goal in a short term. Therefore, suspicious nodes have extremely synchronized behavior patterns within a short time, because they are often required to perform similar tasks together such as payments in specific stores.
%	\item \textit{rare behavior}: The connectivity patterns among high suspicious nodes are very different from the majority. Usually, the density of subgraphs composed of highly suspicious nodes is significantly higher than other parts of the full graph.
%\end{itemize}
Through the analysis of fraudulent activities in historical promotional campaigns on JD.com, it is found that there are usually multiple groups of fraudsters in the same period of promotional campaigns. The fraudsters of different groups are reflected in the 'who buy-from where' graph, that is, there are high-density disjoint subgraphs. 
%We need to find all suspicious groups in a graph.
%%\vspace{-8pt}
%\subsection{Problem Definition}\label{sec_problem_def}
%%\vspace{-2pt}
Based on the 'who buy-from where' graph, the target of the fraud detection problem is not only finding the densest subgraph but also extracting all other unexpected dense subgraphs. In this paper, we formulate the fraud detection as finding $k$-disjoint subgraphs that maximize the sum of densities and propose a simple strategy to select the parameter $k$ automatically and stably. Here, we formally define our problem as follows:

\noindent \textbf{Problem Definition}:
Given a 'who buy-from where' graph $G=(\mathcal{U}\cup \mathcal{V},\mathcal{E})$, the fraud detection problem is to find a group of vertex subsets $S = \{S_1,...,S_{\hat{k}}\}$ and $S_i\subseteq \mathcal{U}\cup\mathcal{V}$. $\hat{k}$ is the selected value of the parameter $k$, which is determined by the property of the graph $G$. 

Here, we make use of the \textit{density score} to measure the density instead of the average density degree.  

\noindent \textbf{Definition 2}
(Density Score): 
Let the density score $\phi(G)$ to be $$\frac{1}{|\mathcal{U}|+|\mathcal{V}|}\left(\sum_{j\in \mathcal{V}}\frac{1}{\log(d_j+c)}\right)$$
where $d_j$ denotes the node degree of the $j_{th}$ merchant, and $c$ is a small constant to prevent the denominator from becoming zero.

%The definition of $\phi(G)$ is introduced by  first. Compared with the average density degrees, the \textit{density score} will detect a fraudulent block without failure if it contains more edges .
According to \cite{hooi2016fraudar}, given a graph $G$, its density can be effectively measured with the \textit{density score}. The reason for penalizing high-degree merchant nodes in \textit{density score} is to keep the detection effective, even in the face of camouflage. The more detailed explanation can reference to \cite{hooi2016fraudar} as well.

Based on the graph \textit{density score} measure, the objective function of the proposed fraud detection problem can be represented as :
%\vspace{-5pt}
\begin{equation}\label{equ:k-disjoint}
	\max\sum_{i=1}^{\hat{k}} \phi(G(S_i))\;\; s.t. S_l\cap S_m = \emptyset\;\;l\neq m\;\;l,m \in \{1, 2, \cdots, \hat{k}\}. 
	%	\vspace{-3pt}
\end{equation}
Here, $G(S_i)$ represents a subgraph which is composed of the vertex subset $S_i$. It is not hard to prove that Equ.~\ref{equ:k-disjoint} is a NP-hard problem \cite{balalau2015finding}.

%\vspace{-8pt}
\section{Proposed Method}\label{sec:method}
%\vspace{-4pt}
In this section, we introduce the proposed model {\our}. At the very beginning, we introduce three structural sampling methods along with analysis. After that, the fraud detection algorithm {\ouralgorithm} will be proposed which can be conducted for all sampled graphs in parallel. Finally, we describe the ensemble approach {\our} from a holistic perspective. 
%\vspace{-8pt}
\subsection{Sampling methods for bipartite graph}\label{sec:bagging}
%\vspace{-2pt}
Most of the real-world graphs are still too large to efficiently acquire, store and process. In order to facilitate the development and testing of systems in network domains, it is often necessary to sample smaller subgraphs from the larger network structure \cite{leskovec2006sampling}. In e-commerce business platforms, the amount of transaction data is normally huge, and the response time for fraud detection is extremely demanding. We make use of sampling methods to decompose the fraud detection problem in a large scale graph into multiple smaller scale graphs which can be solved simultaneously through the parallel implementation. 

There are two different sampling strategies: node sampling and edge sampling. In classic node sampling, nodes are chosen independently and uniformly at random from the original graph for inclusion in sampled graphs. All edges among the sampled nodes are added to form the sampled graph. But for a bipartite graph $G = (\mathcal{U}\cup \mathcal{V},\mathcal{E})$, $\mathcal{U}$ is not as the same type as $\mathcal{V}$, thus we consider two different sampling strategies further: \textit{one-side node sampling} and \textit{two-side node sampling}. \textit{Edge sampling} focuses on the selection of edges rather than nodes to build sampled subgraphs. Thus, the node selection step in edge sampling algorithm proceeds by just sampling edges, and including both nodes when a particular edge is sampled. We next introduce these three sampling choices for bipartite graphs along with analysis in detail.

\subsubsection{Random Edge Sampling (RES)}\label{sec:edge_bagging}
\textit{RES} is the most intuitive and simplest choice of sampling methods in a bipartite graph. The random edge sampling can be conducted with the following steps:
\begin{enumerate} [leftmargin=*,label=\textbf{\arabic*}.]
	\item Select a random set of edges $\mathcal{E}_s$ from
	$G = (\mathcal{U}\cup \mathcal{V},\mathcal{E})$ with the sample ratio $\mathit{S} = |\mathcal{E}^i_s|/|\mathcal{E}|$.
	Collecting nodes connected by edges in $\mathcal{E}^i_s$ into sets $\mathcal{U}^i_s$ and $\mathcal{V}^i_s$ separately.
	\item Based on $\mathcal{E}^i_s$, $\mathcal{U}^i_s$ and $\mathcal{V}^i_s$, we can construct the random sampled subgraph $G_s^i = (\mathcal{U}^i_s\cup\mathcal{V}^i_s,\mathcal{E}^i_s)$.
	
\end{enumerate}

% Because the algorithm {\ouralgorithm} mentioned in Section~\ref{sec:algorithm} takes $\mathcal{O}(|\mathcal{E}|\log|\mathcal{V}|)$ time, 
It's obvious that RES can sample subgraphs evenly and control the time consumption of each subgraph is similar. Based on Lemma 1, RES will sample nodes with larger degrees at a higher rate. What's more, from the view of the spectral, RES is likely to select dense components from the original bipartite graph with the rise of sample ratio $\mathit{S}$. Based on the problem we face, we know that such dense components have a greater likelihood of containing fraudulent nodes. At the same time, the sparse part will not be considered in the process, thus the method prunes low-risk nodes and reduces the hypothetical space during the procedure of ensemble construction.

\subsubsection{One-side Node sampling (ONS)}\label{sec:one_side_bagging}
A notable difference between unipartite graphs and bipartite graphs is the node type. Nodes constituting a unipartite graph only belong to one type, but a bipartite graph has two types of nodes which can be seen as two sides of the graph. Obviously, \textit{ONS} is a straightforward way to process the sampling of a bipartite graph. \textit{ONS} can be operated as the following steps:

\begin{enumerate} [leftmargin=*,label=\textbf{\arabic*}.]
	\item Construct the adjacency matrix $W \in \mathbb{R}^{|\mathcal{U}|\times|\mathcal{V}|}$ for the original bipartite graph $G = (\mathcal{U}\cup \mathcal{V},\mathcal{E})$. 
	\item Determine the node type to sample. Here, assuming $\mathcal{U}$. 
	\item According to the sample ratio $\mathit{S} = |\mathcal{U}^i_s|/|\mathcal{U}|$, sampling $|\mathcal{U}^i_s|$ rows from $W$ to form the subgraph adjacency matrix $W^i_s$, 
	\item Construct the random sampled subgraph $G^i_s$ based on $\mathcal{W}^i_s$. 
	
\end{enumerate}

From the steps above, we can find that determining which side nodes to sample is a non-trivial option for \textit{ONS}. Here, we have summed up some paradigms through practice.
\begin{itemize}[leftmargin=*]
	
	\item \textit{Task-oriented principle}: 
	First of all, we insist that the decision should depend on the problem to solve. For instance, if to deal with the link prediction or closeness measurement problems for one-side nodes, we should sample the target side nodes in order to measure the closeness with the support of global and complete information from the other side. Meanwhile, with the rise of sampled subgraphs, no node pair will be missed. However, when facing problems relating to dense subgraph detection, it's another kind of situation where we should take other properties into consideration like the specific topology. We analyze the situation in the next bullet point. 
	
	\item \textit{Retain topology}: 
	We still take the problem relating to dense subgraph detection as an example. Here, we have a bipartite graph with $\mathit{D}_{avg}(\mathcal{V})\gg\mathit{D}_{avg}(\mathcal{U})$ where $\mathit{D}_{avg}(\mathcal{U})$ is the average degree of $\mathcal{U}$. In response to the task requirement, we care about retaining the topology of dense components from the original graph after sampling. Obviously, if we sample $\mathcal{U}$, the sampled subgraph will become a subgraph of uniform distribution, especially when $\mathit{D}_{avg}(\mathcal{U}) \sim 1$. In contrast, sampling $\mathcal{V}$ can retain dense topology from the original graph effectively. Because once $\mathit{v}\in \mathcal{V}$ with a high degree is sampled, the dense components can be extracted. 
	% In conclusion, the retention of topology will affect the bagging strategy, and the requirement of topology is decided by algorithm task at the same time.
\end{itemize}

ONS can conquer disparate structures more effectively. Considering dense subgraphs problem, it's more effective to select denser subgraphs from relatively dense ones after sampling, and subgraphs that are not very significant to be selected with the global setting become distinct on sampled graphs. 
% The sampling result has strong relationship with the distribution of node degree which will lead to the sampled subgraphs with large difference in size, despite we set the sample ratio $\mathit{S}$ in advance. For example, the node $\mathit{u}$ with $\mathit{D}(u)=\mathit{D}_{total}(\mathcal{U})/2$ once is sampled, the graph size will be larger than half of the original graph. In this way, it's hard to control the size of subgraphs with the sample ratio $\mathit{S}$. The direct consequence of this situation is the huge variance of subgraphs' time consumption which will affect the efficiency of parallel implementation to some extent.   
%\vspace{-5pt}
\subsubsection{Two-sides Node Sampling (TNS)}\label{sec:two_side_bagging}
% After introducing the one-side node bagging for the bipartite graph, i
It's hard to avoid thinking about what if we conduct sampling to both sides of the bipartite graph. In fact, it's a fairly intuitive idea. From the view of the adjacency matrix, it is equivalent to sampling both rows and columns of $W$ and using the cross-section of these rows and columns to construct the adjacency matrix $W_s^i$. Hence the detailed sampling steps are similar to \textit{ONS}.

% and are described as following:
% \begin{enumerate}
% \item Construct the $|\mathcal{U}|\times|\mathcal{V}|$ adjacency matrix $W$ for the original bipartite graph $G$. 
% \item Set the sample ratio $\mathit{S} = |\mathcal{U}_r|/|\mathcal{U}|=|\mathcal{V}_r|/|\mathcal{V}|$, and sample $|\mathcal{U}_r|$ rows and $|\mathcal{V}_r|$ columns from $W$. Take the cross cells between rows and columns to construct the subgraph adjacency matrix $W_r^k$,
% \item Apply the algorithm {\ouralgorithm} to $W_r^k$ and save the detected node sets $\mathcal{U}_f^k$ and $\mathcal{V}_f^k$.
% \item To repeat step (2)-(3) in sequence or in parallel for $\mathit{N}$ times as setting. Get sets of all detected nodes $\mathcal{U}_f$ and $\mathcal{V}_f$. 
% \item Apply the aggregation method to $\mathcal{U}_f$ or $\mathcal{V}_f$ and output the detected nodes.
% \end{enumerate}

Here we need to remind a point to better understand the number of sampled subgraphs $\mathit{N}$ and the sample ratio $\mathit{S}$. Because we sample both sides of the bipartite graph, the sampled graph is smaller than RES or ONS with the same sample ratio $\mathit{S}$. In fact. when $\mathit{S}=0.1 $, the size of a subgraph from random edge sampling is $10\%$ of the original one, but the subgraph sampled from two sides is only near $\mathit{S}^2$ of the original graph. So in practice, we usually need to enlarge the sampling ratio $\mathit{S}$ or increase the number of samples $\mathit{N}$ to ensure the effectiveness of the two-sides sample. However, considering the parallel implementation of {\our}, with the rise of $\mathit{N}$, more computing resources need to be occupied at the same time which may be unacceptable in real-world scenes.
Besides, it's easy to understand that \textit{TNS} will preserve the structure of the graph more finely and make it difficult to preserve specific topology as needed like \textit{ONS}, which is an aspect need to be considered.

% \subsection{Bound of Edge Bagging Ensemble}
% \begin{theorem}
% Theorem 1. Suppose we have oracle access to the edge set of the input graph with the ability to sample an edge uniformly at random. There exists an algorithm for the densest subgraph problem running in time $O(m+n)$, which gives a ($0.5-\epsilon$)-approximate solution, with probability $1-\frac{1}{n}$.
% \end{theorem}

% Algorithm 2 samples $O(\frac{n\delta\log(l)}{\gamma\epsilon^2})$ edges of the input graph and runs Algorithm.~\ref{alg:FDdenseblockdetectionet} on the sampled graph. Interestingly, the following Theorem shows Algorithm 2 is an ($\alpha-\epsilon$)-approximation algorithm for P on G.

% \begin{theorem}
% Theorem 2. Let G be an arbitrary graph G, and let Algorithm 1 be an $\alpha$-approximation algorithm for P. With probability $1-e^{-\delta}$, Algorithm 2 is an ($\alpha -\epsilon$)-approximation algorithm for P on G, using $O(\frac{n\delta\log(l)}{\gamma\epsilon^2})$ space.
% \end{theorem}

% \begin{lemma}
% Let G be the input graph, and let Alg be an $\alpha$-approximation algorithm, Let $G_i=(U_i\cup W_i,\epsilon)$ be the sampled graph by algorithm 2 and let $p = \frac{c}{m}$. We have
% \begin{equation}
% Pr(\alpha opt(G,K)-f_G(Alg(H)))\geq 6\epsilon opt(G)) 
% \le 6\exp (\log(|Sol_G^k|)-\frac{p\epsilon^2 opt(G)}{12f_k})
% \end{equation}
% \end{lemma}
%\vspace{-8pt}

\subsection{Fraud DETection({\ouralgorithm}) algorithm based on Heuristic}\label{sec:Heuristic}
%\vspace{-2pt}
The sampling methods are able to decompose the large scale graph into multiple smaller scale graphs. Now, we propose {\ouralgorithm} to complete fraud detections for each sampled subgraph. As we mentioned in Section~\ref{sec_problem_def}, the fraud detection task in our realistic e-commerce scene is a disjoint case. We start from considering one natural heuristic for the disjoint case: at each step, we compute the densest subgraph $G(S_i)=(S_i,\mathcal{E}_i)$ in the current graph $G$. Formally, to achieve $G(S_i)$, nodes in $G$ which result in the highest value of \textit{density score} $\phi$ defined in Definition 2 will be repeatedly obtained. Then we remove edges in previously detected subgraphs from the current graph $G$. Iteratively searching the dense subgraph until we find the $\hat{k}_{th}$ subgraph or the current graph contains no edges. 
%\vspace{-10pt}
\begin{algorithm}[t]
	%	\large
	\caption{{\ouralgorithm} for Equ.~\ref{equ:k-disjoint}}
	%	\large
	\label{alg:denseblockdetection}
	\KwData{Bipartite Graph $G=(\mathcal{U}\cup \mathcal{V},\mathcal{E})$, density metric $\phi$}
	\KwResult{$S_d =(\mathcal{U}_d\cup \mathcal{V}_d)$}
	$\mathcal{U}_d\leftarrow \emptyset$, $\mathcal{V}_d\leftarrow \emptyset$\;
	\Repeat{$\arg\min_{i}\Delta^2\phi(G(S_i))$ and $G\neq\emptyset$}{
		$n \leftarrow |\mathcal{U}|+|\mathcal{V}|$\;
		$H_n \leftarrow G$\;
		\For{$i=n:2$}
		{find minimal degree node $m$ in $H_i$\;
			$H_{i-1} \leftarrow H_i - \{m\}$\;
		}
		$G(S_i) = (\mathcal{U}_i\cup \mathcal{V}_i,\mathcal{E}_i) \leftarrow \arg\max_{H_i \in \{H_2, \dots, H_n\}}  \phi(H_{i})$\;
		$\mathcal{U}_d \leftarrow \mathcal{U}_d \cup \mathcal{U}_i$\;
		$\mathcal{V}_d \leftarrow \mathcal{V}_d \cup \mathcal{V}_i$\;
		remove $\mathcal{E}_i$ from G\;}
\end{algorithm}
%\vspace{-5pt}

How to determine $\hat{k}$ is a very important issue. We insist that $\hat{k}$ should be a parameter that exists objectively depending on how many dense subgraphs exist. Therefore, when detecting dense subgraphs, we have to truncate the detection process effectively instead of setting a number or the more the better. About truncating effective dense components, the basic idea behind partitioning methods, such as k-means clustering \cite{jain2010data}, is to define clusters such that the total intra-cluster variation or total WSS (within-cluster sum of square) is minimized. The total WSS measures the compactness of the clustering, and we want it to be as small as possible. The Elbow method \cite{ketchen1996application,thorndike1953belongs} treats the total WSS as a function of the number of the clusters: one should choose a number of clusters so that the total WSS doesn't improve significantly while adding another cluster. In this paper, we employ a similar idea to select the $\hat{k}$. We see the total density measure $\sum_i^{\hat{k}}\phi(G(S_i))$ as a function of $\hat{k}$: one should choose a number so that $\sum_i^{\hat{k}+1}\phi(G(S_i))$ doesn't improve much better. In other words, $\Delta^2\phi(G(S_i))$ which is the second-order finite difference of $\phi(G(S_i))$ can be utilized to measure the improvement of $\sum_i^{\hat{k}}\phi(G(S_i))$, and $min(\Delta^2\phi(G(S_i)))$ represents the density score $\phi$ suddenly decreases. Therefore, we can define \textit{Truncating Point} as:

\noindent \textbf{Definition 3}
(Truncating Point): 
$$\hat{k} = \arg\min_i\Delta^2\phi(G(S_i))$$

\begin{figure}[h]
	%	\vspace{20pt}
	\centering
	\begin{minipage}[l]{1\columnwidth}
		\centering
		\includegraphics[width=1\textwidth]{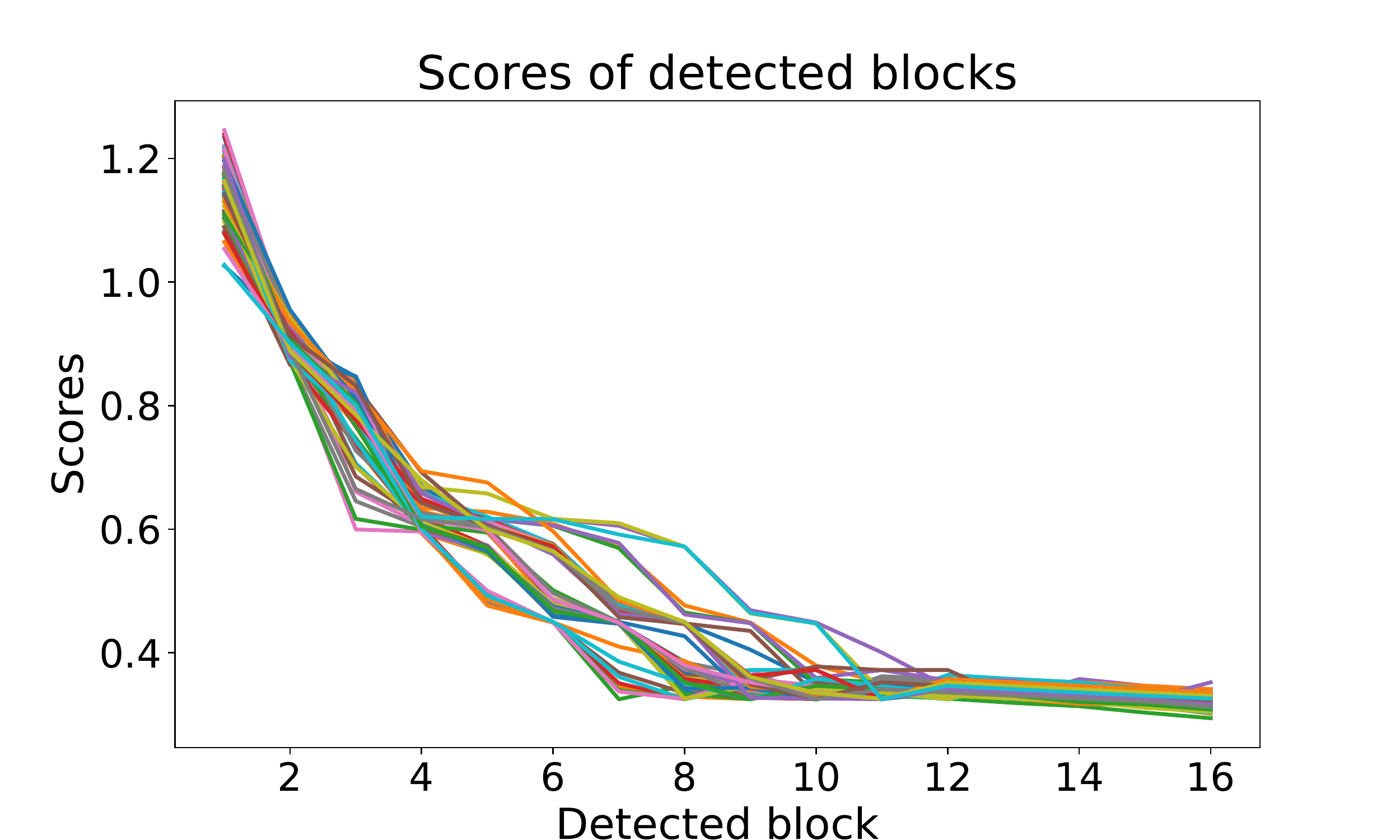}     
	\end{minipage}
	%	\vspace{-10pt}
	\caption{Scores for each detected block.}\label{fig:score}
	%	\vspace{-15pt}
\end{figure}

We plot the curve of $\phi(G(S_i))$ for multiple sampled graphs as shown in Figure~\ref{fig:score}, and one line represents one sampled graph. It can demonstrate this method is able to find a reasonable hyperparameter $\hat{k}$ because all curves are monotonically decreasing and achieve a similar low score finally which means detected subgraphs are meaningless after the truncating point $\hat{k}$. In the experiments, we will compare the performances between a fixed $k$ and the optimal $\hat{k}$.

By using minimal heap \cite{hooi2016fraudar}, each update can be performed in $O(\log(|\mathcal{U}|+|\mathcal{V}|))$ time, totaling $O(\hat{k}|\mathcal{E}|\log(|\mathcal{U}|+|\mathcal{V}|))$ time because we need $|\mathcal{E}|$ updates to node degrees and repeat $\hat{k}$ times. We show {\ouralgorithm} in Algorithm~\ref{alg:denseblockdetection}.
Typically, the number $\hat{k}$ of groups varies from few to few tens. The required number of components $\hat{k}$ is often expected to increase with the graph size. However, based on sampling methods, the large scale graph can be decomposed into multiple smaller scale graphs. Besides, the algorithm {\ouralgorithm} is able to run on each sampled subgraphs in parallel. After describing graph sampling methods and the fraud detection algorithm in each subgraph, we will introduce our proposed model {\our} from a holistic view.
%\vspace{-10pt}
%-----------------------------------------------------------------
\begin{figure*}[t]
	\vspace{30pt}
	\centering
	\begin{minipage}[l]{2\columnwidth}
		\centering
		\includegraphics[width=1\textwidth]{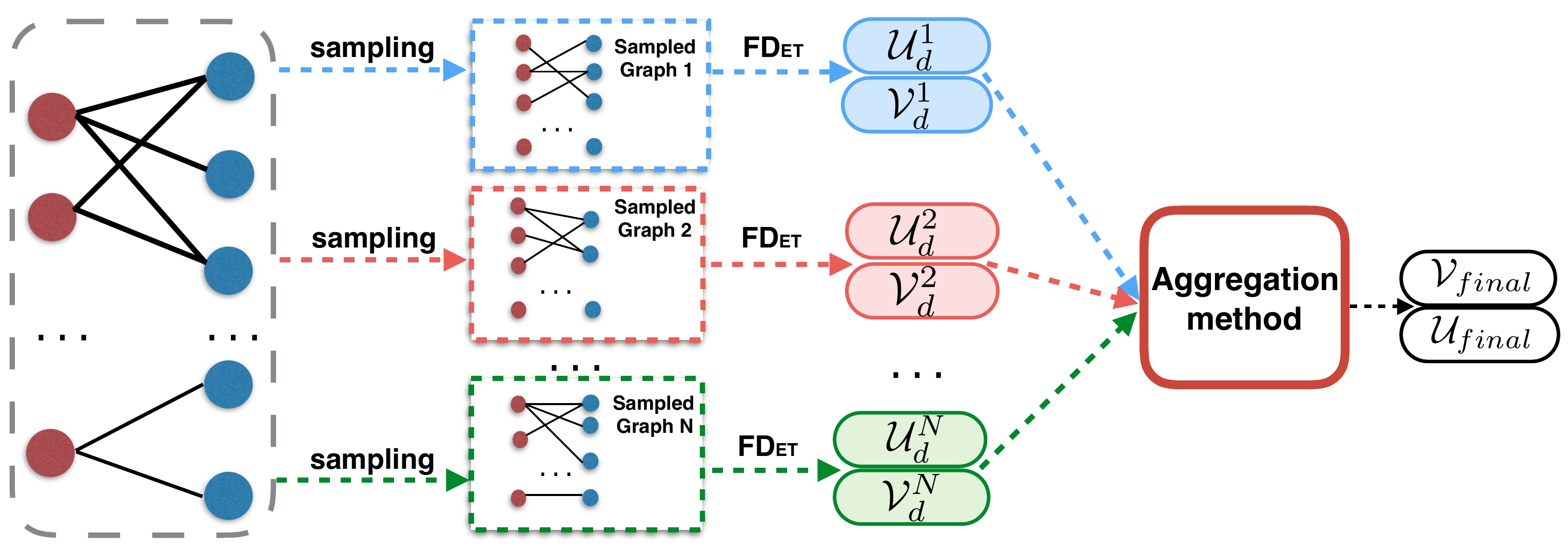}
	\end{minipage}
	%	\vspace{-10pt}
	\caption{The Structure of {\our}}\label{fig:framework}
	%	\vspace{-10pt}
\end{figure*}
%------------------------------------------
\subsection{Ensemble based Framework {\our}}\label{sec:ensemfdet}
{\our} is a general fraud detection method and can be applied to detect various fraud behaviors in real-world scenes. The structure of {\our} is shown in Figure~\ref{fig:framework}. Meanwhile, the pseudo-code of {\our} is available in Algorithm~\ref{alg:Framwork}. 

In {\our}, the process before applying the aggregation method can be implemented in parallel. After sampling, we will apply {\ouralgorithm} to all sampled graphs simultaneously with the multicore environment. The parallel property of {\our} is very meaningful and practical because there are high demands for running time in the real-world where graphs are very huge in size. Then {\our} applies the aggregation method to the results from all sampled graphs to get the final fraud detection output. In experiments, we choose the majority voting aggregation method as defined:
%\vspace{-4pt}

\noindent \textbf{Definition 4}
(Majority Voting Aggregation Method ($MVA$)): The majority voting aggregation method can be described by this equation: 
$$\mathit{H}(u) =
\begin{cases}
accept       & \quad \text{if } 
\left(\sum_{i=1}^{N} \mathit{h}_i(u)\right) \geq T\\
reject       & \quad \text{otherwise.}
\end{cases}$$
where $\mathit{h}_i(u)$ is the number of the vote-catching of the node $u$ in sampled subgraph $G_s^i$, and the parameter $\mathit{T}$ is a threshold.

%\vspace{-4pt}
The parameter $\mathit{T}$ normally is determined by task requirements and the preference for detection results.
Although the heuristic algorithm {\ouralgorithm} is able to solve the fraud detection based on bipartite graphs with near-linear scalability, it is still unable to control the size of dense subgraphs which may limit its practicality. The threshold $\mathit{T}$ is able to handle the quantity of detected suspicious nodes, and we will analyze the impact of $\mathit{T}$ in Section~\ref{sec:parameter}. In fact, the aggregation methods are flexible and can be set as the one suitable for the specific requirement. 
% In the business scene, normally the graph is very huge in size, and the business unit also has a high demand for running and update time. Therefore, t

%\begin{algorithm}
%	\caption{{\our}}
%	\label{alg:Framwork}
%	\begin{algorithmic}[1]
%	  \Require  
%	    Bipartite Graph$G=(\mathcal{U}\cup \mathcal{V},\mathcal{E})$;\ Density score metric $\phi$;
%		Sample Method $\mathcal{M}$;\ Number of sampled graphs $\mathit{N}$;\ Sample ratio $\mathit{S}$;\ Voting threshold $\mathit{T}$
%	  \Ensure 
%	    Detected sets of fraud nodes $\mathcal{U}_{final}$ and $\mathcal{V}_{final}$
%	  
%    \end{algorithmic}
%\end{algorithm}
%\vspace{-5pt}
\begin{algorithm}
	\caption{{\our}}
	\small
	\label{alg:Framwork}
	\KwIn{Bipartite Graph $G=(\mathcal{U}\cup \mathcal{V},\mathcal{E})$;\ Density score metric $\phi$;\ Sample Method $\mathcal{M}$;\ Number of sampled graphs $\mathit{N}$;\ Sample ratio $\mathit{S}$;\ Voting threshold $\mathit{T}$;\ Majority Voting Aggregation Method $MVA$;}
	\KwOut{Detected sets of fraud nodes $\mathcal{U}_{final}$ and $\mathcal{V}_{final}$}
	$\mathcal{U}_d\leftarrow \emptyset$, $\mathcal{V}_d\leftarrow \emptyset$\;
	\Begin(\textbf{run in parallel}){
		
		\For{$i \in \{1,2,\dots,\mathit{N}\}$}{
			Apply Sample Method $\mathcal{M}$ to $G$ with $\mathit{S}$ and get $G_s^i$\;
			Apply {\ouralgorithm} to $G_s^i$ and get $S_d^i=(\mathcal{U}_d^i\cup \mathcal{V}_d^i)$\;
			$\mathcal{U}_d \leftarrow \mathcal{U}_d \cup \mathcal{U}_d^i$\;
			$\mathcal{V}_d \leftarrow \mathcal{V}_d \cup \mathcal{V}_d^i$\;
		}
	}
	$\mathcal{U}_{final}\leftarrow \emptyset$, $\mathcal{V}_{final}\leftarrow \emptyset$\;
	Apply $MVA$ to $\mathcal{U}_d$, $\mathcal{V}_d$ with the voting threshold $\mathit{T}$\;
	\For{$u \in \mathcal{U}_d$}{
		\If{$\mathit{H}(u) = accept$}{
			$\mathcal{U}_{final}\leftarrow \mathcal{U}_{final} \cup u$}
	}
	\For{$v \in \mathcal{V}_d$}{
		\If{$\mathit{H}(v) = accept$}{
			$\mathcal{V}_{final}\leftarrow \mathcal{V}_{final} \cup v$}
	}
	\Return $\mathcal{U}_{final},\mathcal{V}_{final}$;
\end{algorithm}

%\vspace{-10pt}
\section{Experiments}\label{sec:experiment}
%\vspace{-4pt}

To demonstrate the effectiveness, stability, and scalability of {\our}, extensive experiments are conducted in three datasets that come from real transaction data on \textbf{JD.com}. In this section, we will describe datasets in detail at first. Then the experimental settings, including experimental setup, evaluation metrics, and comparison methods, will be introduced. Finally, we will display the experimental results together with the parameters impact analysis.
%\vspace{-8pt} 
\subsection{Datasets Description}
\begin{table}[h]
	%	\vspace{-10pt}
	\caption{Statistics of datasets}
	%	\normalsize
	\label{tab:datastat}
	%	\vspace{-10pt}
	\begin{tabular}{p{12mm}ccccl}
		\toprule
		Dataset\# &Node:PIN & Fraud PIN & Node:Merchant & Edge\\
		\midrule
		\texttt{1} & 454,925 & 24,247 & 226,585 & 1,023,846 \\
		\texttt{2} & 2,194,325 & 16,035 & 120,867 & 2,790,517\\
		\texttt{3} & 4,332,696 & 101,702 & 556,634 & 7,997,696\\
		\bottomrule
	\end{tabular}
	%	\vspace{-15pt}  
\end{table}
Datasets we used in experiments are based on real transaction data on \textbf{JD.com}, one of the world's largest e-commerce business platforms. A set of policies, rules, and models will determine levels of risk of all transactions on \textbf{JD.com}, and a subset of them with relatively high risk will be sent for additional manual checking. A team consisting of experienced experts in manual checking will review those transactions carefully and determine if they should be rejected. Once transactions are rejected, the \textit{PIN} of users participating transactions will be marked as dangerous and be recorded in the Blacklist. The Blacklist is used as the ground-truth for evaluating fraud detection algorithms. 
% A more detailed description of the business scenario and data generation process can refer to Section~\ref{sec:fraud_modes}. 
After cleaning up the original mass transaction data, we get \textit{PIN-Merchant} bipartite graph ('who buy-from where' graph) which describes the trading relationship between users and merchants. For these three datasets, we also have three corresponding blacklists that contain dangerous \textit{PIN} of users as the ground-truth. 
% What needs to be explained here is that t
These three datasets actually have to be independent of each other, because they are collected from different time periods, and the business scene we face is extremely time-sensitive. For example, one \textit{PIN} appears in three datasets, but only in one dataset, it is marked as black which may be due to the theft of accounts. And later because of some operations like appeals. the \textit{PIN} can be removed from the blacklist. The key statistical data describing datasets can be found in Table~\ref{tab:datastat}. We hold the view that in response to this kind of practical problem, experiments conducted on the datasets from the real-world are the most valuable and convincing.

%\vspace{-5pt}
\subsection{Experimental Settings}
%\vspace{-2pt}
\subsubsection{Experimental Setup and Metrics}
In the experiments, {\our} can be divided into three steps and conduct them step by step just like the description in Section~\ref{sec:ensemfdet}.
% At first, we sample the dataset into multiple subgraphs with introduced sampling methods in Section~\ref{sec:bagging}. The parameters $\mathit{S}$ and $\mathit{N}$ will determine the scale and the number of sampled bipartite graphs.
%% , and we will analyze the impact of these parameters in Section~\ref{sec:parameter}. 
% Then {\ouralgorithm} will be applied into every sampled bipartite graphs simultaneously with the support of the multicore environment. Next, the aggregation method will work on the outputs from all subgraphs to get final result relating to the threshold $\mathit{T}$. 
The key parameters are summarized in Table~\ref{tab:parameters}
% which will determine the performance of {\our}
, and extensive experiments with different combinations of parameters are conducted in order to discover interesting impacts.   
\begin{table}[h]
	%	\scriptsize
	\caption{Parameters used in experiments}
	%	\normalsize
	\label{tab:parameters}
	%	\vspace{-8pt}
	\centering
	\begin{tabular}{p{15mm}<{\centering}|cl}
		\toprule
		Parameters &Descriptions \\
		\midrule
		$\mathit{N}$ & Number of sampled graph  \\
		$\mathit{S}$ & Sample ratio\\
		$\mathit{T}$ & Voting threshold in aggregation method\\
		$\mathit{R}$ & The repetition rate $\mathit{R}=\mathit{S}\times\mathit{N}$\\
		\bottomrule
	\end{tabular}
	%	\vspace{-10pt}
\end{table}
% \subsubsection{Metrics}

The main goal of our experiments is to compare the quality and quantity of the detected fraud nodes, thus we can choose conventional evaluation metrics to measure the performance. 

The methods we test in experiments can all output detected fraud nodes, thus we can use F1, Recall, Precision as evaluation metrics. It should be noted that normally Accuracy in fraud detection problems seems not very significant, because the proportion of fraud samples is quite low.

\subsubsection{Comparison Methods}

The methods used in experiments are listed as following:
\begin{itemize}[leftmargin=*]
	\item \textbf{\our}: {\our} is the model proposed in this paper. We aim at demonstrating the practicability and stability of {\our} and the advantage of the ensemble framework in time consumption without loss of detection performance in our experiments.
	\item \textbf{{\ourfixedk}}: In {\ourfixedk}, we fix the number of detected blocks $k$ instead of truncating the detecting process automatically which is described in Section~\ref{sec:Heuristic}. The method is used to verify the effectiveness of the truncation. 
	\item \textbf{{\spoken}}:  By spectral relaxation, {\spoken} \cite{prakash2010eigenspokes} is able to find the densest density regions with SVD method. 
	% Compared with {\our}, 
	{\spoken} is not parameter-free to make sure how many components should be used to estimate the suspicious nodes. So in our experiments, the number of components is set to $25$ as same as the paper described. 
	\item \textbf{\Fbox}:  {\Fbox}\cite{shah2014spotting} analyzes the reconstruction error and shows attacks of small enough scale cannot be efficiently detected in the top-$k$ SVD components.
	% Thus this method can be regarded as a complementing way to SVD based methods.
	{\Fbox} needs to set the parameter $k$ which is a determinant factor of the reconstruction error.
	% Beyond the top-$k$ SVD components, 
	\item \textbf{\Fraudar}: {\Fraudar} \cite{hooi2016fraudar} tries to find an unexpected dense subgraph which commonly contains high suspicious users.
	
\end{itemize}

Based on fair considerations, as an unsupervised learning method, we will not compare with supervised learning methods, such as some feature-based models and emerging GNN-based models~\cite{wang2019fdgars,ZCXJXL18,park2019fraud}.
%\vspace{-7pt}
\subsection{Experiment Result}
%\vspace{-2pt}
\subsubsection{Evaluation on Comparison Methods}\label{sec:comparison_method_evaluation}

\begin{figure}[h]
	\vspace{30pt}
	\centering
	\subfigure[Dataset \#1]{ \label{fig:comparison_small_dataset}
		\begin{minipage}[l]{1\columnwidth}
			\centering
			\includegraphics[width=1\textwidth]{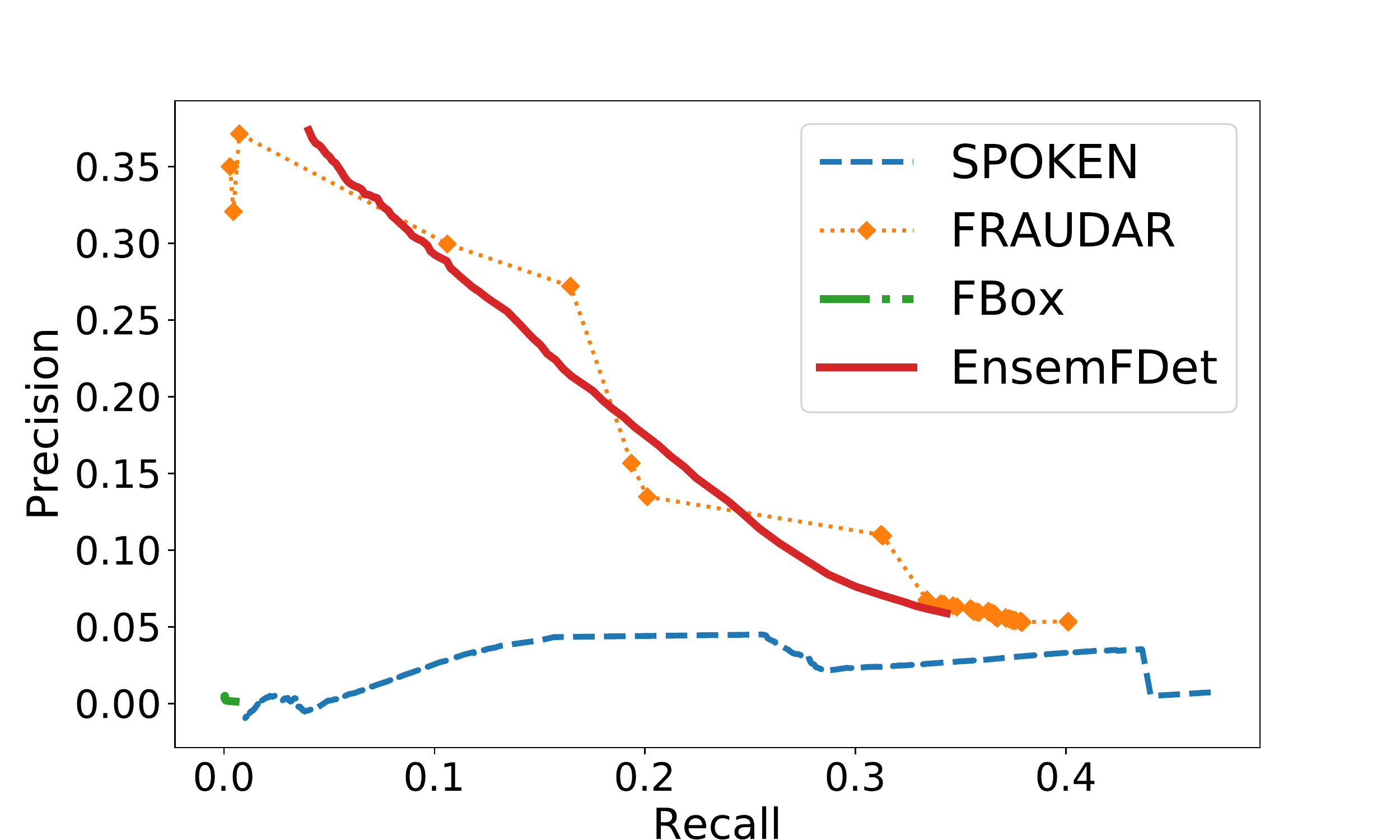}\vspace{5pt}
		\end{minipage}
	}
	\subfigure[Dataset \#2]{\label{fig:comparison_large_dataset}
		\begin{minipage}[l]{1\columnwidth}
			\centering
			\includegraphics[width=1\textwidth]{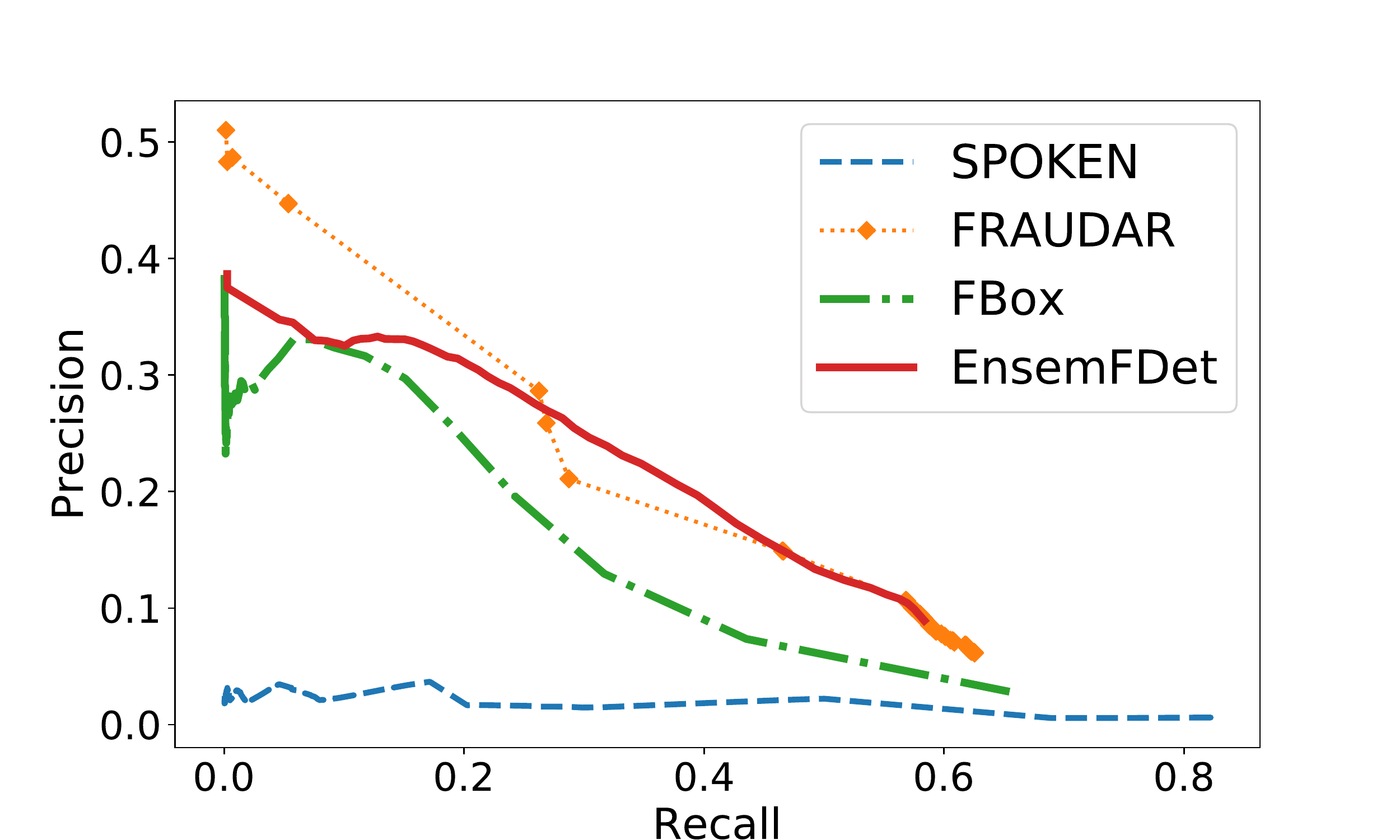}\vspace{5pt}
		\end{minipage}
	}
	\subfigure[Dataset \#3]{\label{fig:comparison_final_dataset}
		\begin{minipage}[l]{1\columnwidth}
			\centering
			\includegraphics[width=1\textwidth]{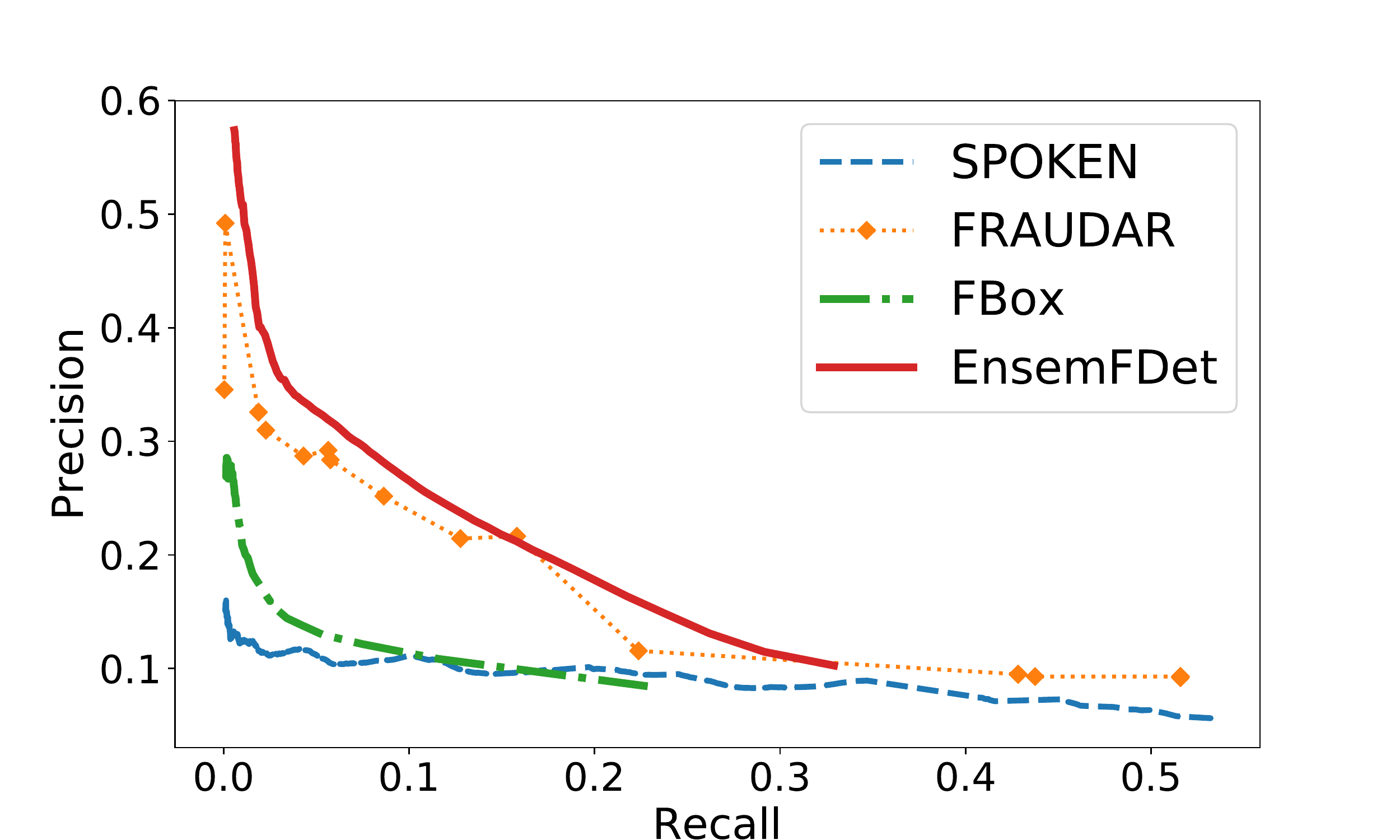}\vspace{5pt}
		\end{minipage}
	}
	%	\vspace{-10pt}
	\caption{Performance comparison of different methods}\label{fig:performance_comparison_method}
	%	\vspace{-8pt}
\end{figure}

\begin{figure*}[h]
	\vspace{20pt}
	\centering
	\subfigure[Dataset \#1]{ \label{fig:Fraudar_ensem_small_dataset}
		\begin{minipage}[l]{0.70\columnwidth}
			\centering
			\includegraphics[width=1\textwidth]{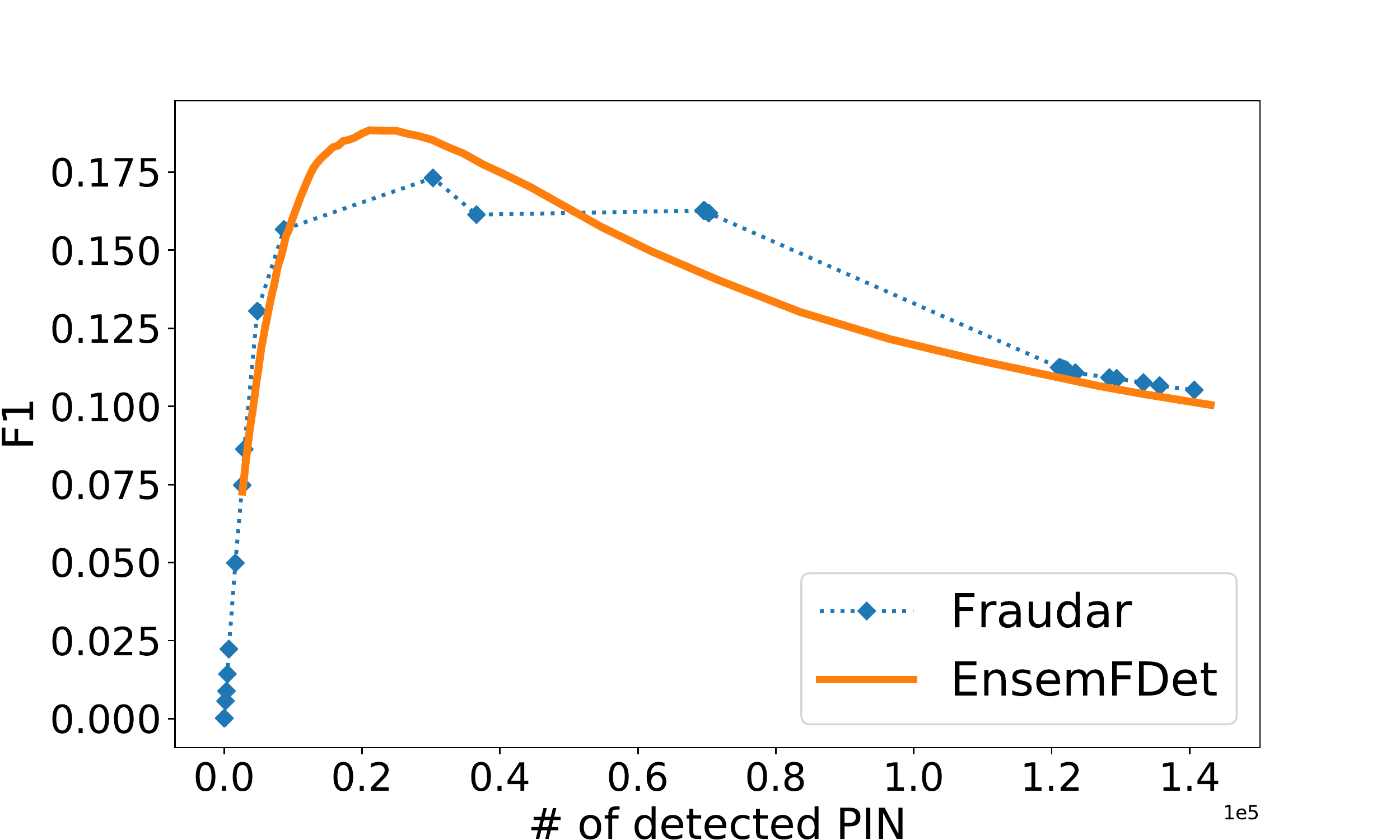}\vspace{3pt}
		\end{minipage}
	}\hspace{-22pt}
	%	\vspace{-5pt} 
	\subfigure[Dataset \#2]{\label{fig:Fraudar_ensem_large_dataset}
		\begin{minipage}[l]{0.70\columnwidth}
			\centering
			\includegraphics[width=1\textwidth]{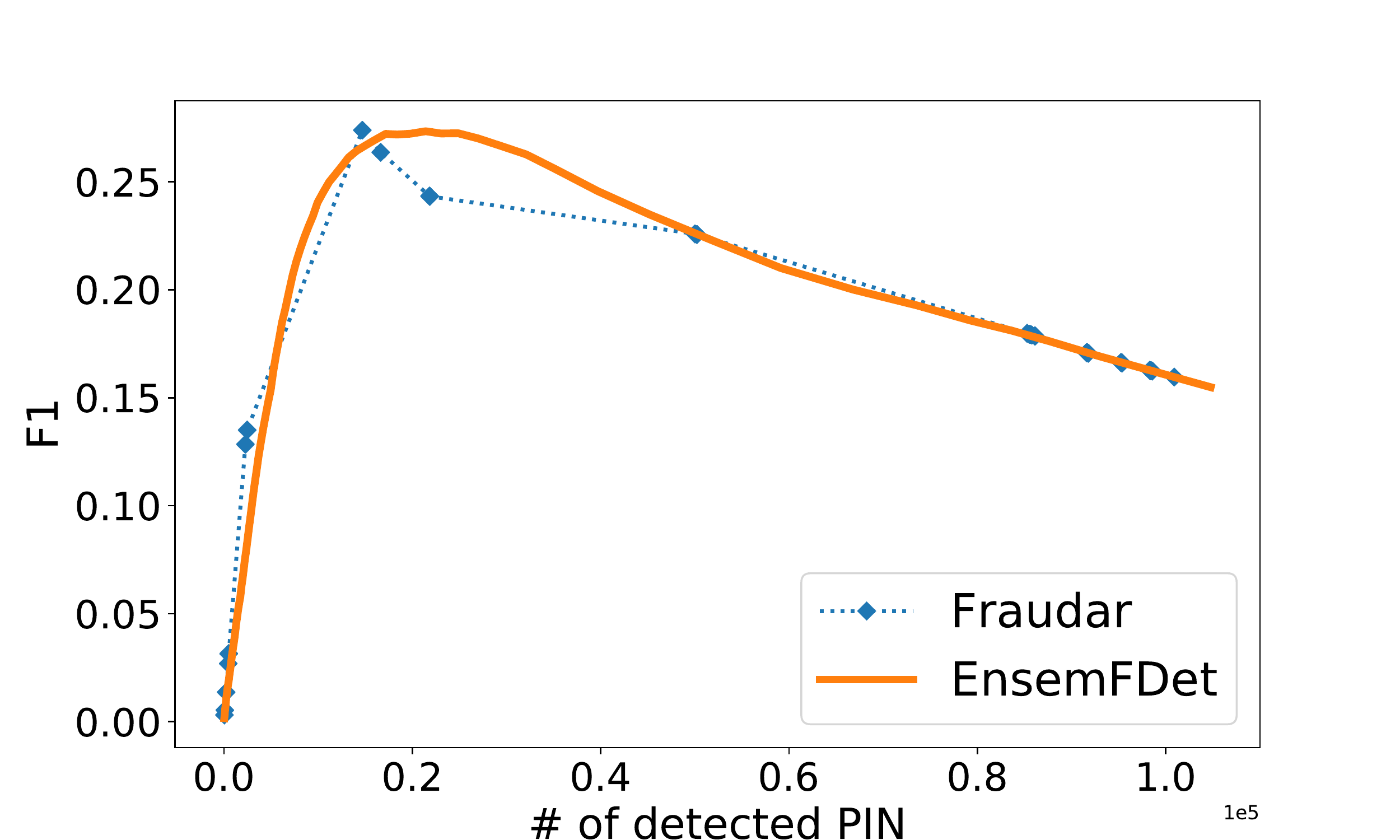}\vspace{3pt}
		\end{minipage}
	}\hspace{-22pt}
	%	\vspace{-5pt}
	\subfigure[Dataset \#3]{\label{fig:Fraudar_ensem_final_dataset}
		\begin{minipage}[l]{0.70\columnwidth}
			\centering
			\includegraphics[width=1\textwidth]{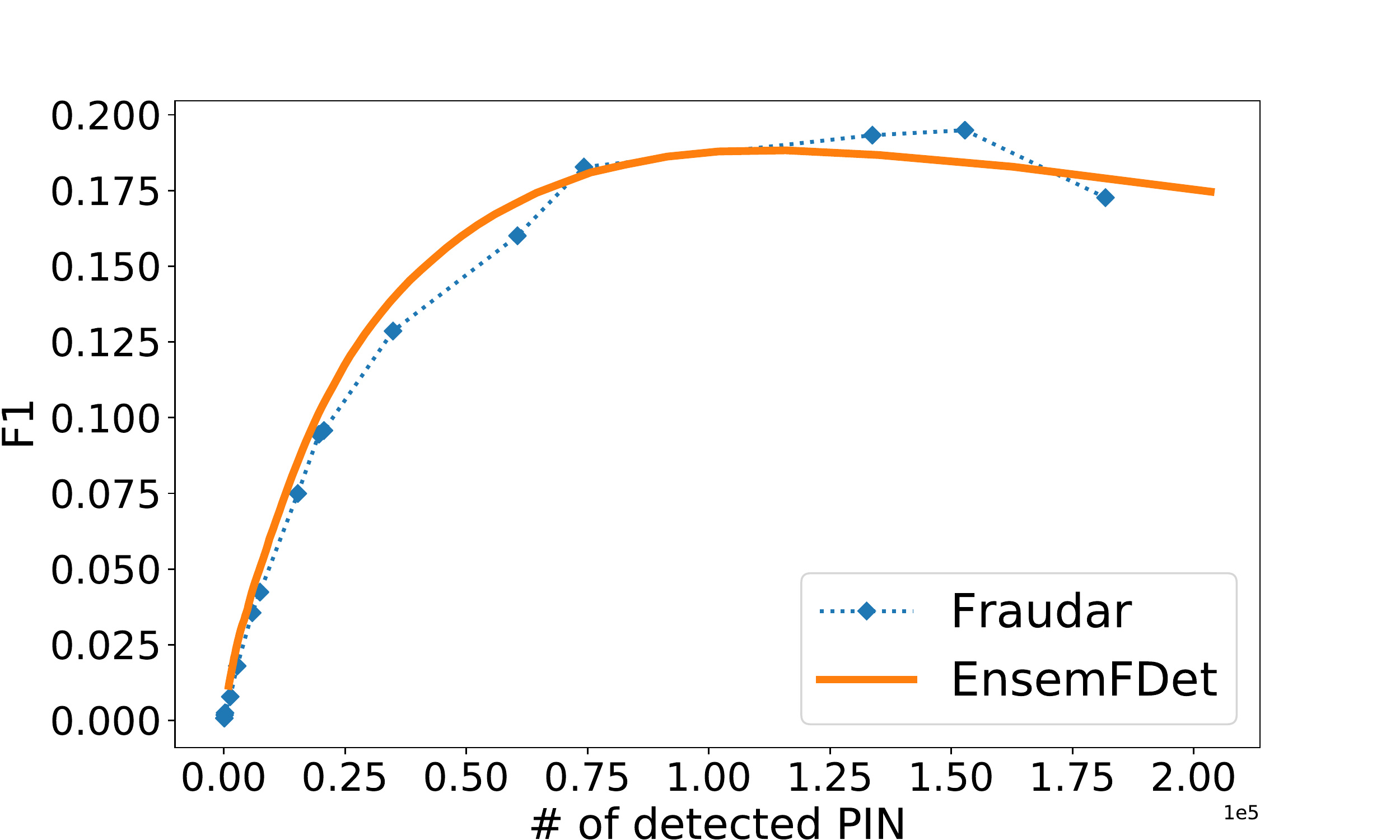}\vspace{3pt}
		\end{minipage}
	}\hspace{-22pt}
	%	\vspace{-5pt}
	\subfigure[Dataset \#1]{\label{fig:Fraudar_ensem_small_dataset_precision}
		\begin{minipage}[l]{0.70\columnwidth}
			\centering
			\includegraphics[width=1\textwidth]{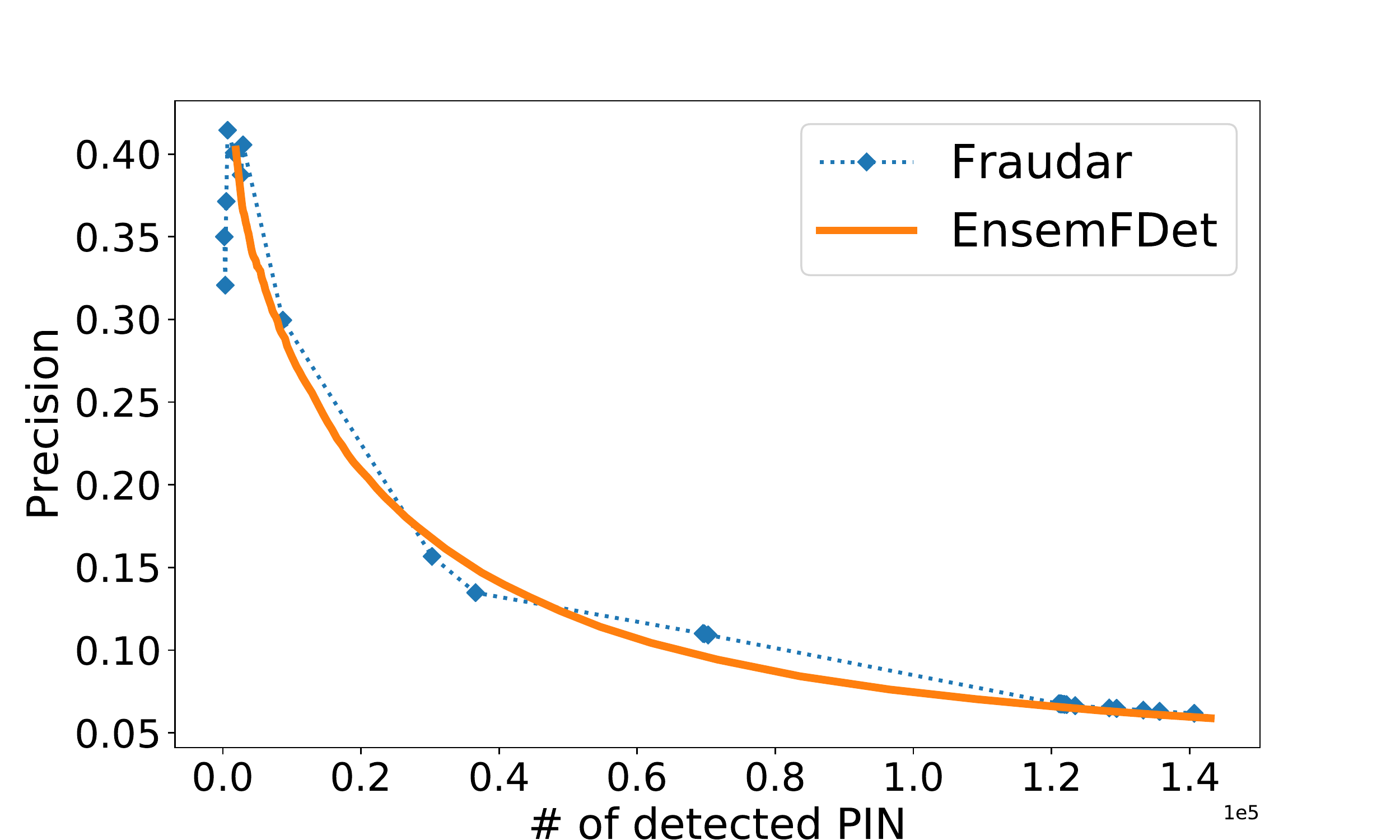}\vspace{3pt}
		\end{minipage}
	}\hspace{-22pt}
	\subfigure[Dataset \#2]{\label{fig:Fraudar_ensem_large_dataset_precision}
		\begin{minipage}[l]{0.70\columnwidth}
			\centering
			\includegraphics[width=1\textwidth]{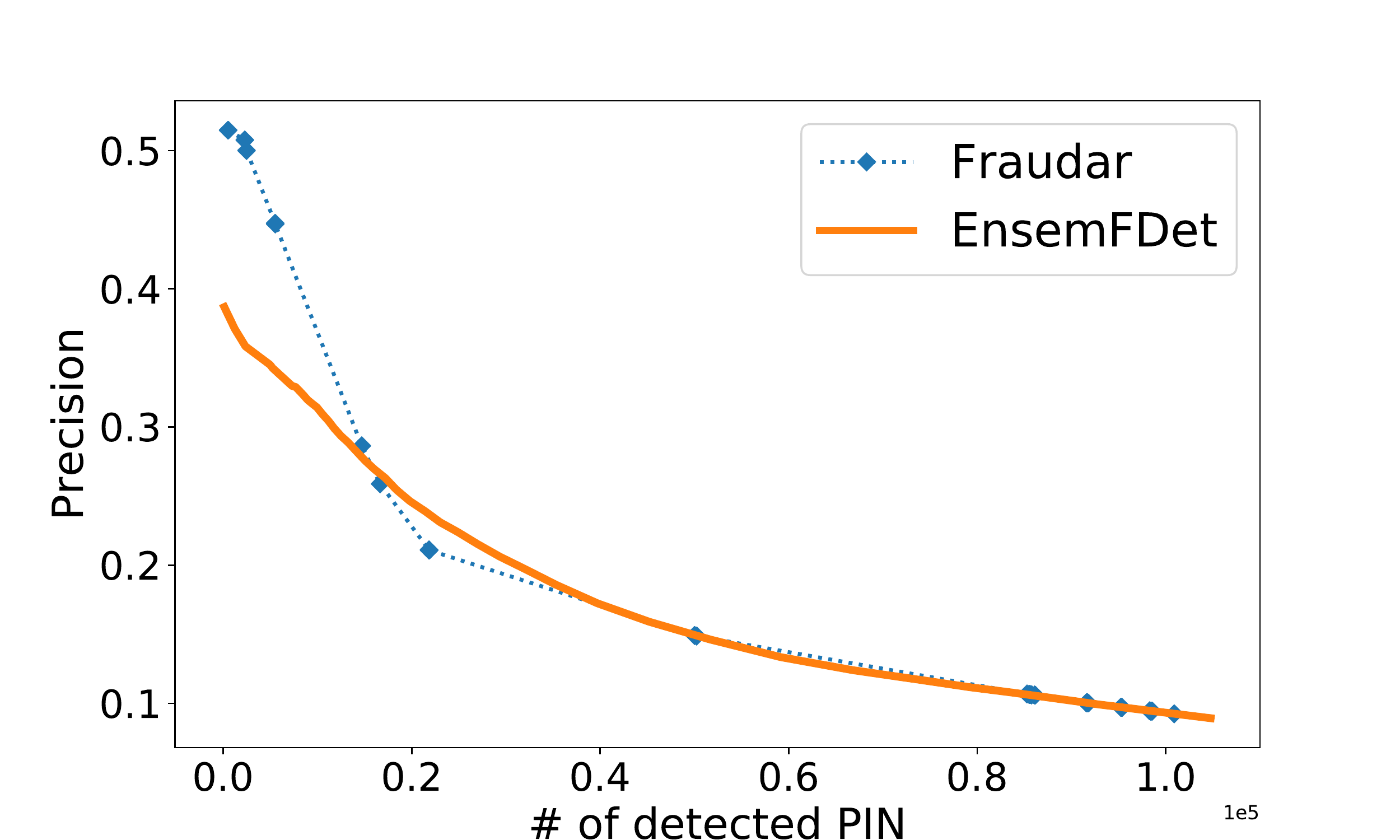}\vspace{3pt}
		\end{minipage}
	}\hspace{-22pt}
	\subfigure[Dataset \#3]{\label{fig:Fraudar_ensem_final_dataset_precision}
		\begin{minipage}[l]{0.70\columnwidth}
			\centering
			\includegraphics[width=1\textwidth]{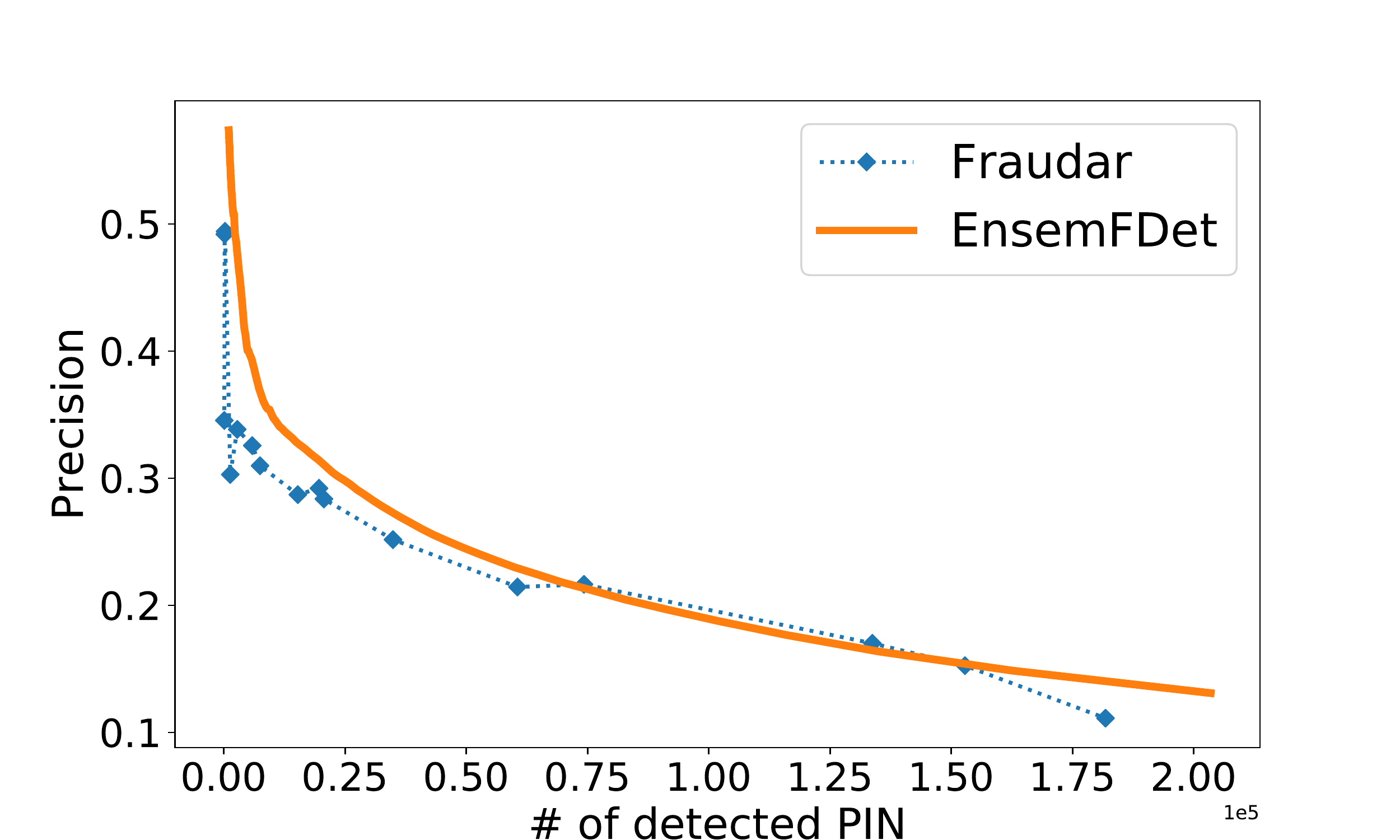}\vspace{3pt}
		\end{minipage}
	}
	%	\vspace{-2pt}
	\caption{Performance and properties Analysis between {\our} and {\Fraudar}}\label{fig:analysis_fraudar_ensem}
	%	\vspace{-10pt}
\end{figure*}

The experimental results of all comparison methods in three datasets are shown in Figure~\ref{fig:performance_comparison_method}. We can find that methods based on SVD, including {\spoken} and {\Fbox}, are not able to keep a stable performance in different datasets. Especially for {\Fbox}, almost completely invalidated on No.1 Dataset: Precision and Recall are close to $0$, but it works on the other two datasets. It's obvious that {\Fraudar} and {\our} can get better performance in all datasets. Because the performance of {\Fraudar} can not be represented with a continuous curve, we use diamond points to represent experimental results of {\Fraudar}.  
%------------------------------------------
%------------------------------------------
Overall, {\our} has close performance compared with {\Fraudar}. let's make more detailed comparison from Figure~\ref{fig:analysis_fraudar_ensem} where we set $\mathit{S}$ as $0.1$ and $\mathit{N}$ as $80$. The number of detected nodes is used in the x-axis of Figure~\ref{fig:analysis_fraudar_ensem}, because in response to the algorithms' task, the performance has a comparative value when {\our} and {\Fraudar} detect the equivalent fraud nodes. From Figure~\ref{fig:Fraudar_ensem_small_dataset}-~\ref{fig:Fraudar_ensem_final_dataset}, we can find {\our} and {\Fraudar} have similar performance in F1. However, what we are trying to express from these figures is not only a similar performance but also the practicability of {\our}. {\our} shows a smooth curve on Figure~\ref{fig:analysis_fraudar_ensem}, because the numbers of fraud nodes being detected in {\our} are almost continuous with the control of $\mathit{T}$. In comparison, the numbers of detected nodes in {\Fraudar} are marked with diamond points, and it's obvious that their connection is a polyline. This means that the number of nodes in the detected blocks is unstable so that we are unable to control the number of nodes being detected by {\Fraudar} and select one point that best fits our requirements on the curve. Besides, on Figure~\ref{fig:Fraudar_ensem_large_dataset_precision} we can notice that there are several marks of {\Fraudar} have apparent advantages compared with {\our}, but from Figure~\ref{fig:Fraudar_ensem_large_dataset} this kind of advantage disappears when taking F1 into consideration. It reflects Recall is very too low to be used in the real world. And at the next point, {\Fraudar} is going to be weaker than {\our}, which spans almost $20,000$ nodes. Normally, such a huge span is unacceptable in the business, but {\our} can be used through the entire curve which makes it practical in the face of real-world scenes.
%------------------------------------------
%------------------------------------------

In terms of time consumption, we compared the running time of {\our} and {\Fraudar} which are both  heuristic methods. Table~\ref{tab:time_consumption} shows the running time of experiments where $\mathit{S}=0.1$, $\mathit{N}=80$ for {\our}, and $K$ is fixed as $30$ for {\Fraudar}. {\Fraudar} and {\our} both runs near-linearly in the input size which verifies and ensures the scalability. Because {\our} can be parallel and make use of the truncation strategy to decrease the number of detected blocks. {\our} is 10X faster than {\Fraudar}. In theory, $Time({\our}) < \mathit{S}\times Time({\Fraudar})$ which is demonstrated
%\vspace{-2pt}
in our experiments. For the smallest $\mathit{S}=0.01$ we tested, {\our} can be 100x faster due to its parallelism.
\begin{table}[t]
	%	\normalsize
	\vspace{50pt}
	\caption{The comparison of time consumption between {\our} and {\Fraudar} }
	\label{tab:time_consumption}

	\centering
	\begin{tabular}{p{18mm}|cccl}
		\toprule
		\qquad   &Dataset \#1 & Dataset \#2 & Dataset \#3 \\
		\midrule
		{\our} & 74.127 sec & 162.102 sec & 470.508 sec  \\
		{\Fraudar} & 805.533 sec & 2365.659 sec & 5681.591 sec\\
		\bottomrule
	\end{tabular}
	%	\vspace{-20pt}
\end{table}
%\vspace{-5pt}
\subsubsection{Comparison among Sampling Methods }
\begin{figure}[h]
	%	\vspace{30pt}
	\centering
	\begin{minipage}[l]{1\columnwidth}
		\centering
		\includegraphics[width=1\textwidth]{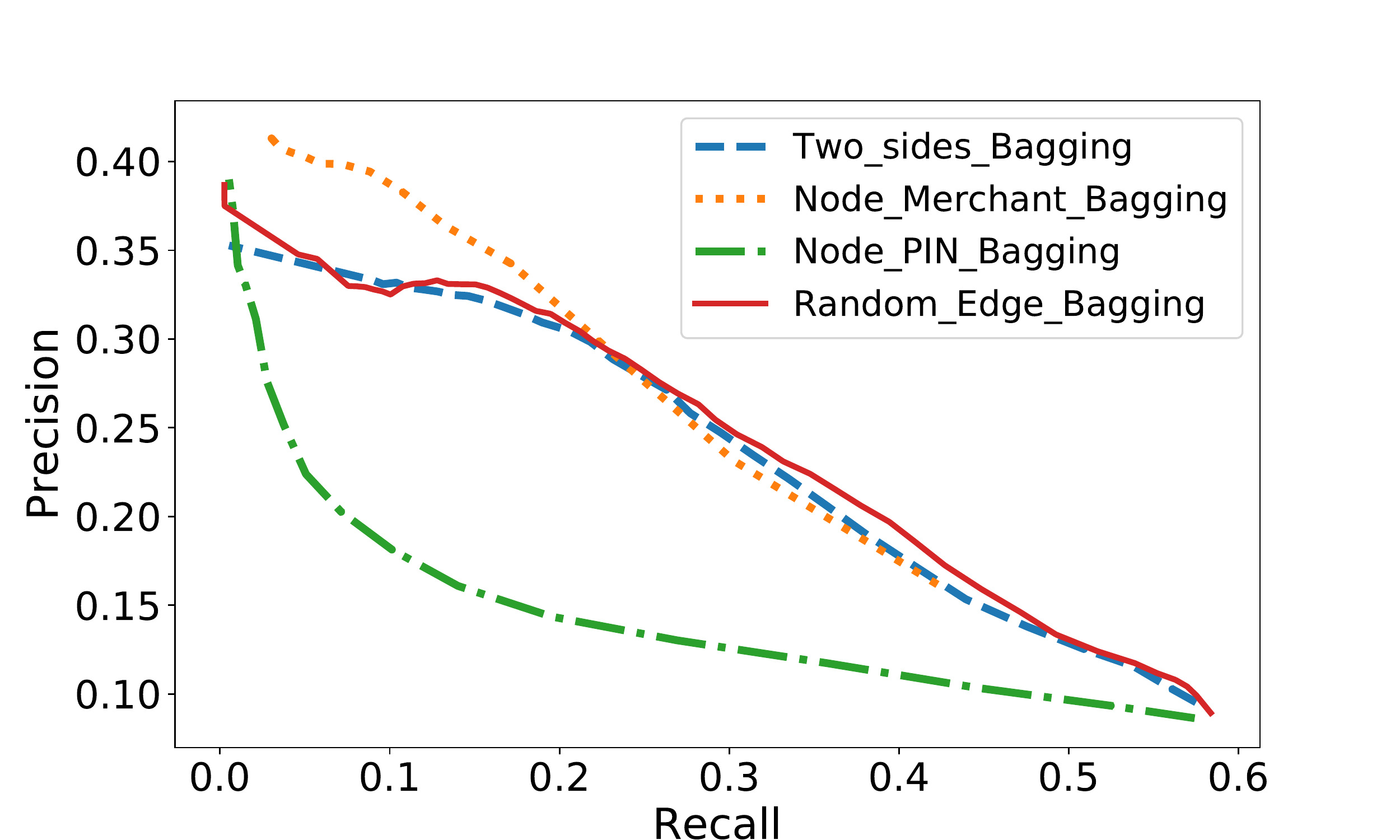}
	\end{minipage}
	%	\vspace{-4pt}
	\caption{Performance comparison among different sampling methods in {\our}.}\label{fig:sample_comparison} 
	%	\vspace{-10pt}
\end{figure}

In Section~\ref{sec:bagging}, we analyzed several sampling methods for a bipartite graph, and here experiments are conducted to demonstrate our strategy and analysis. The Precision-Recall curves in Figure~\ref{fig:sample_comparison} come from the experiments conducted on the No.3 dataset where $\mathit{S}=0.1$ and $\mathit{R}=8$.
%------------------------------------------
%------------------------------------------
% The figure displays the results from four different bagging methods which are noted in the legend of the figure.
At first, from Figure~\ref{fig:sample_comparison} we can find the performance of \textit{Node PIN Bagging} (apply \textit{ONS} as the sampling method in {\our}) is worst, but the curve of \textit{Node Merchant Bagging} is much better. In fact, the performance verifies our analysis in Section~\ref{sec:one_side_bagging}. In this dataset, the graph with $\mathit{D}_{avg}(Merchant)\gg\mathit{D}_{avg}(PIN)$ leads to the failure of retaining the dense topology from the original graph effectively by \textit{Node PIN Bagging}. Conversely, \textit{Node Merchant Bagging} can achieve better performance with the support of keeping critical topology. However, some sampled graphs coming from \textit{Node Merchant Bagging} can be very large in size due to some nodes with a very high degree. Although we set $\mathit{S}$ as $0.1$, the size of some sampled graphs can reach $30\%$ of the original one which results in the higher time-consumption of parallel computation. Under this circumstance, we can discover \textit{Node Merchant Bagging} has better performance. Besides, the similar and stable performance of \textit{Node Merchant Bagging}, \textit{Two-sides Node Bagging} and \textit{Random Edge Bagging} reflects the stability of {\our} to a certain extent.

\subsubsection{Verification of the truncating point}
\begin{figure}[t]
	%	\vspace{30pt}
	\centering
	\begin{minipage}[l]{1\columnwidth}
		\centering
		\includegraphics[width=1\textwidth]{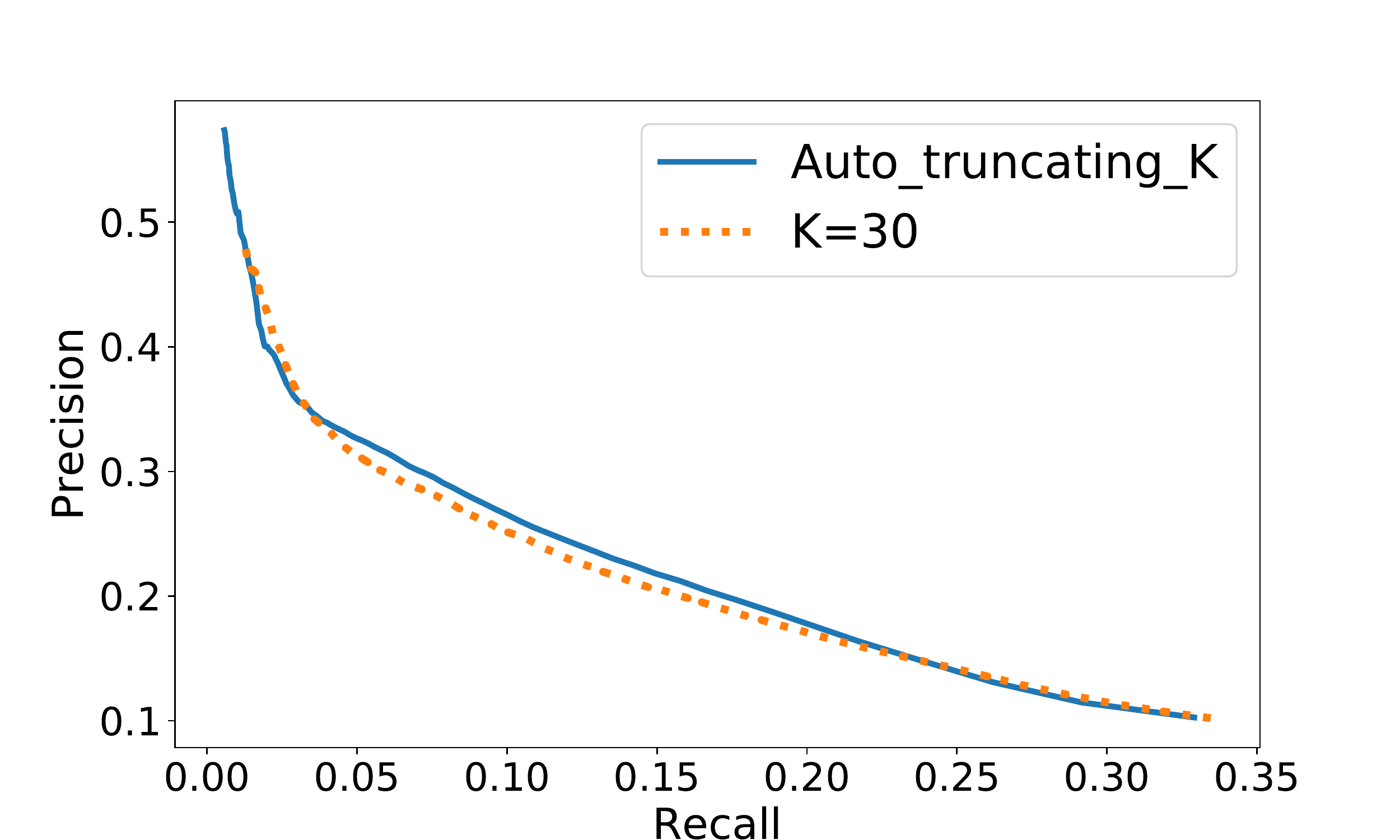}
	\end{minipage}
	%	\vspace{-5pt}
	\caption{Performance comparison between {\our} and {\ourfixedk}.}\label{fig:k_comparison} 
	%	\vspace{-15pt}
\end{figure}

Figure~\ref{fig:k_comparison} displays the results from a comparative experiment between {\our} and {\ourfixedk} where the number of detected blocks $k$ is fixed instead of truncating the detecting process automatically. We set $k=30$ for {\ourfixedk}, and {\our} is based on the truncating point. In experiments, we also record the detected blocks number of {\our}, and all of the records are smaller than $15$. The performance expressed by the Precision-Recall curve shows {\our} can achieve better outcomes than {\ourfixedk}. Although {\ourfixedk} can get higher Recall with the increase of $k$, actually Precision has been close to random selection. This kind of high Recall is meaningless and these blocks after truncation are not of value as we defined before. Therefore, the comparison verifies the effectiveness of the truncation strategy which can even level up the performance in Precision. What's more, the time-consumption has also been greatly reduced, because {\our} only need to detect less than half of $k$ sets for {\ourfixedk} in experiments.
%------------------------------------------
%------------------------------------------
%------------------------------------------
\begin{figure*}[h]
	\vspace{30pt}
	\centering
	\subfigure[Precision-Recall curve]{ \label{fig:N_PR_curve}
		\begin{minipage}[l]{0.49\columnwidth}
			\centering
			\includegraphics[width=1.1\textwidth]{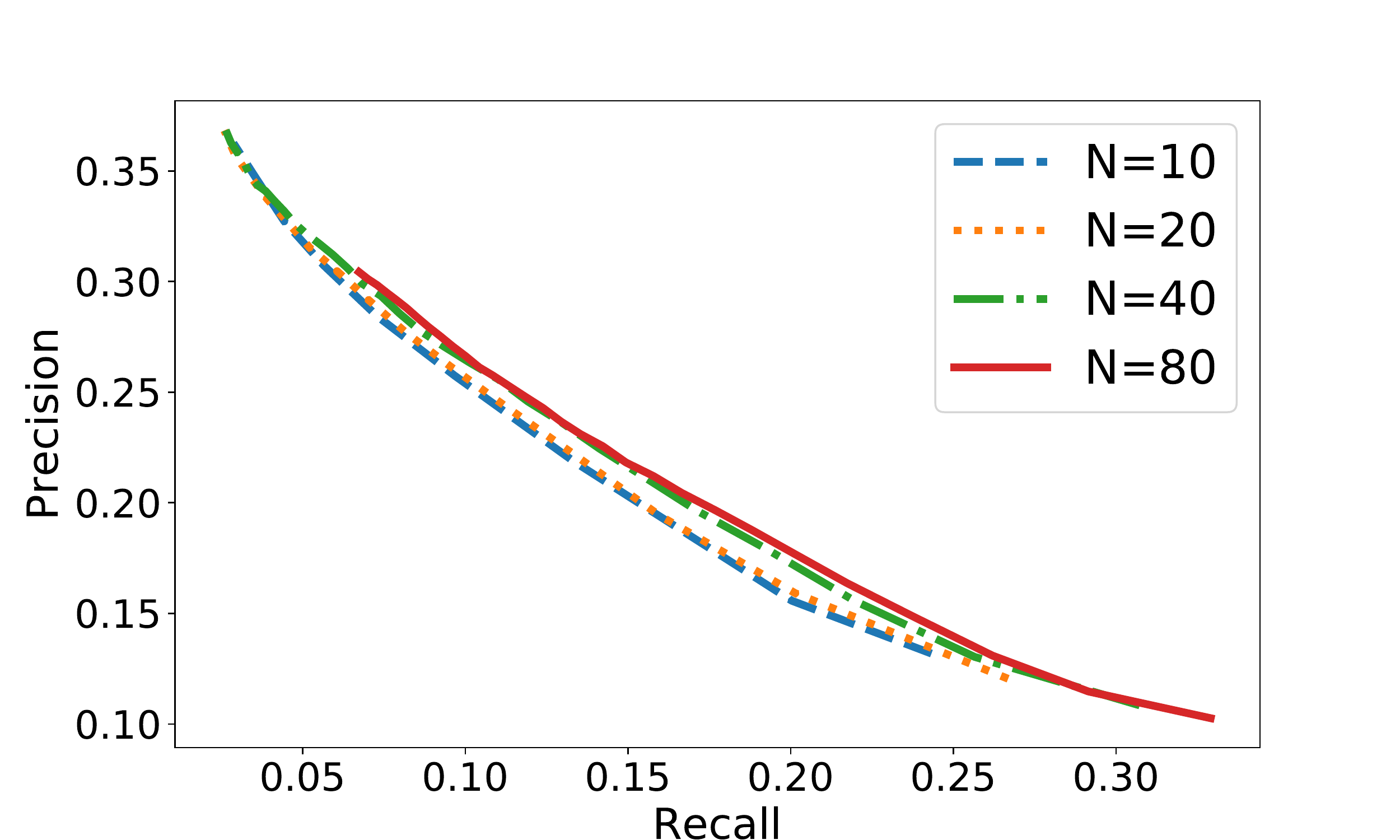}\vspace{5pt}
		\end{minipage}
	}\hspace{-5pt}
	\subfigure[F1]{\label{fig:N_F1}
		\begin{minipage}[l]{0.49\columnwidth}
			\centering
			\includegraphics[width=1.1\textwidth]{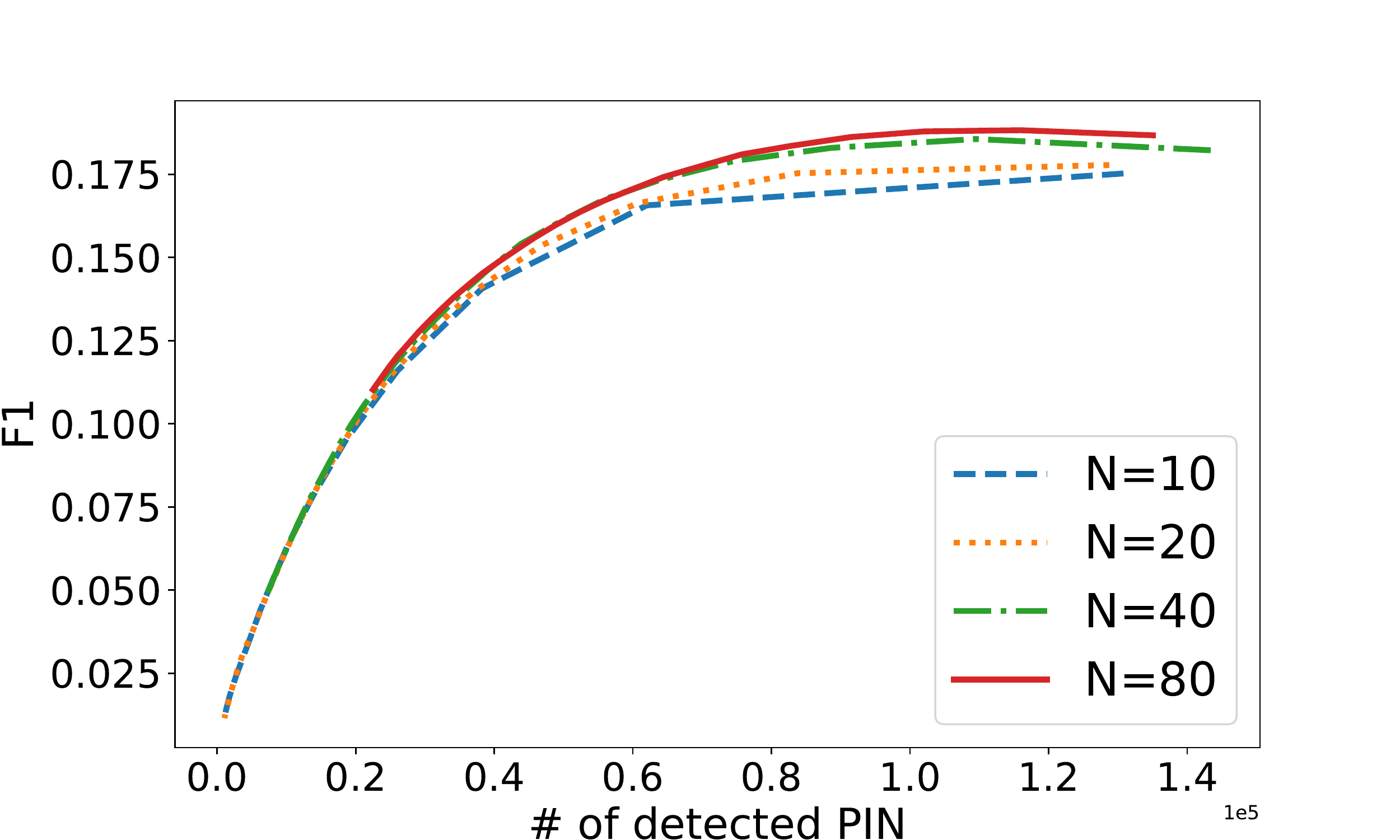}\vspace{5pt}
		\end{minipage}
	}\hspace{-5pt}
	\subfigure[Recall]{\label{fig:N_recall}
		\begin{minipage}[l]{0.49\columnwidth}
			\centering
			\includegraphics[width=1.1\textwidth]{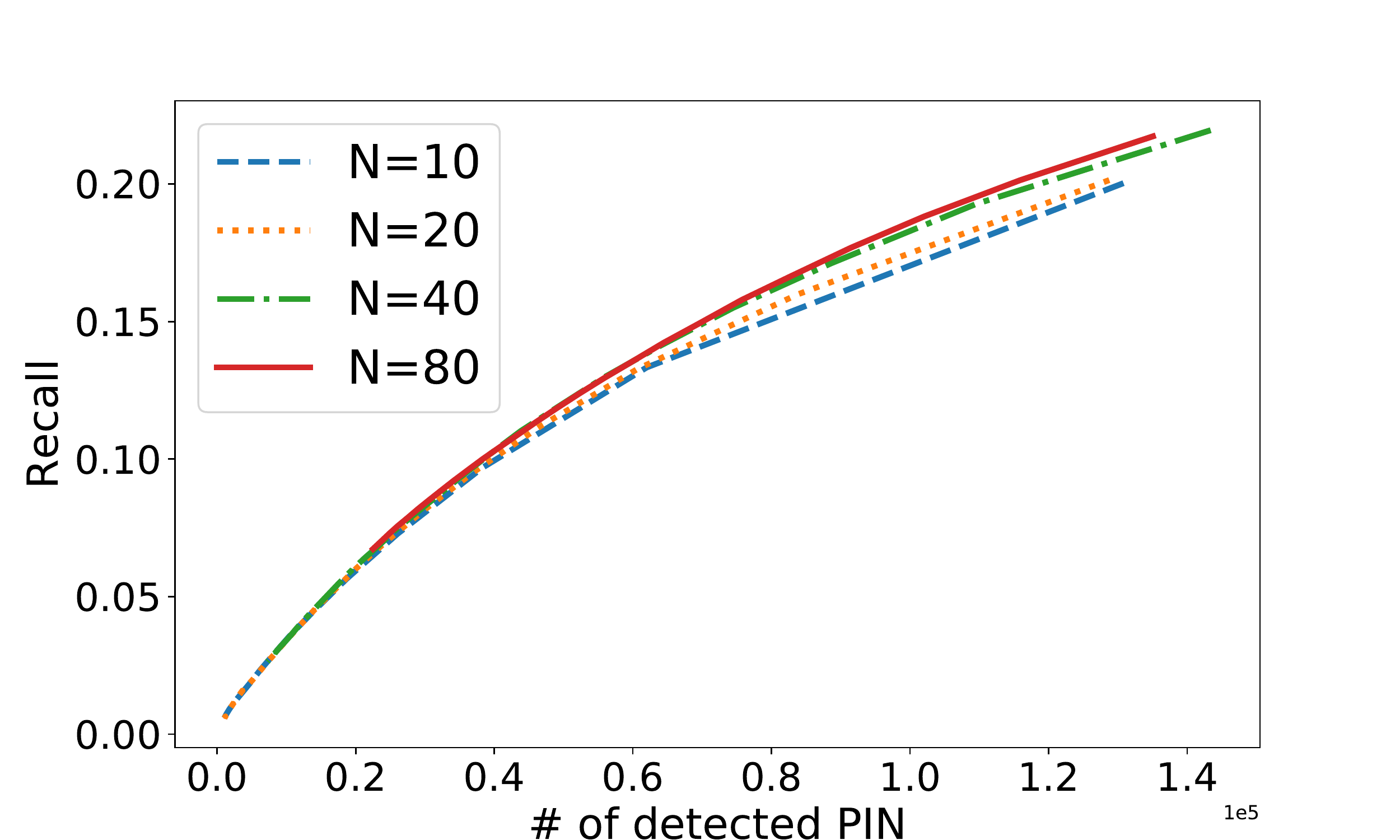}\vspace{5pt}
		\end{minipage}
	}\hspace{-5pt}
	\subfigure[Precision]{\label{fig:N_precision}
		\begin{minipage}[l]{0.48\columnwidth}
			\centering
			\includegraphics[width=1.1\textwidth]{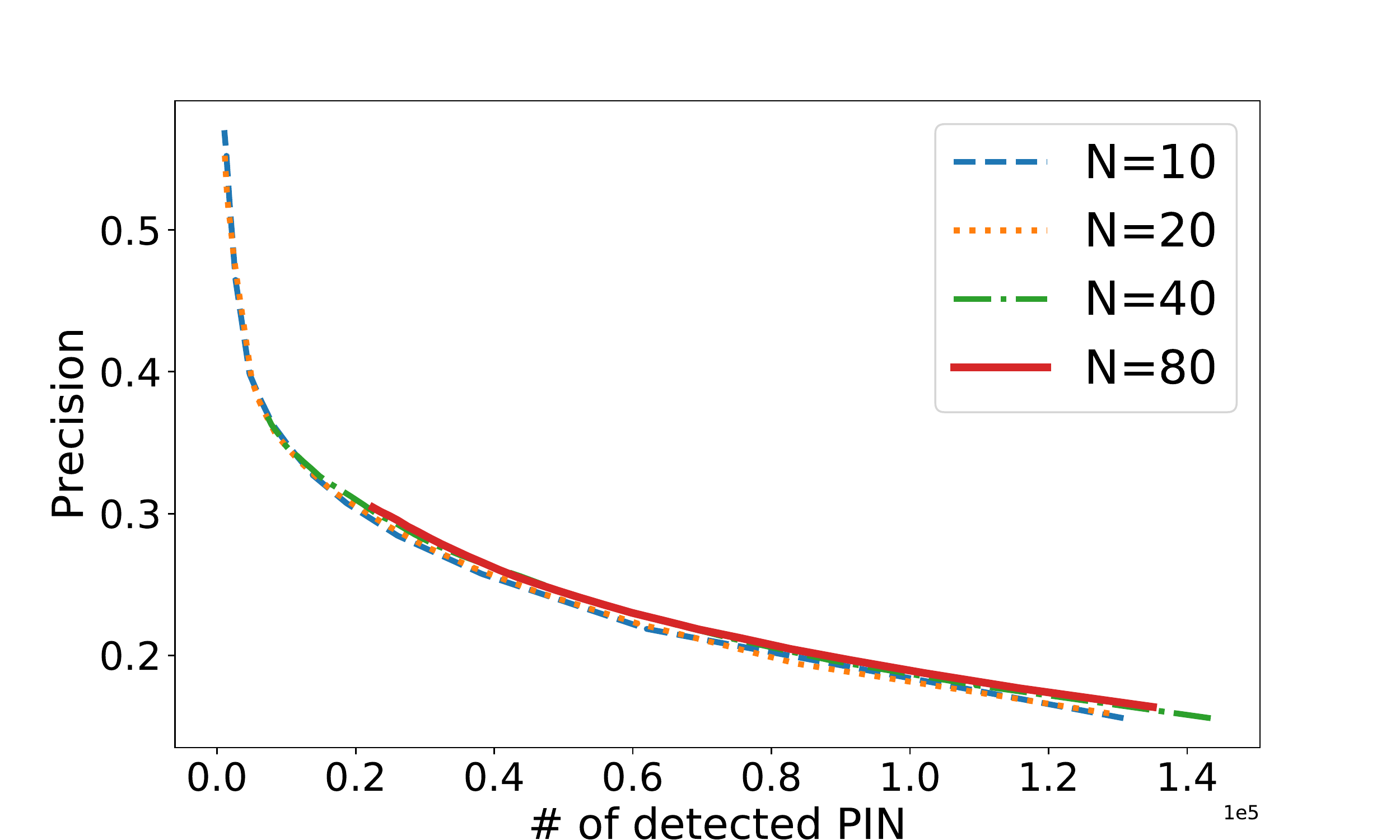}\vspace{5pt}
		\end{minipage}
	}
	%	\vspace{-5pt}
	\caption{Performance Analysis under different $\mathit{N}$ when $\mathit{S}=0.1$}\label{fig:parameter_analysis_N}
	%	\vspace{-5pt}
\end{figure*}
%------------------------------------------
%------------------------------------------
\begin{figure*}[h]
	%	\vspace{-10pt}
	\centering
	\subfigure[Precision-Recall curve]{ \label{fig:S_PR_curve}
		\begin{minipage}[l]{0.49\columnwidth}
			\centering
			\includegraphics[width=1.1\textwidth]{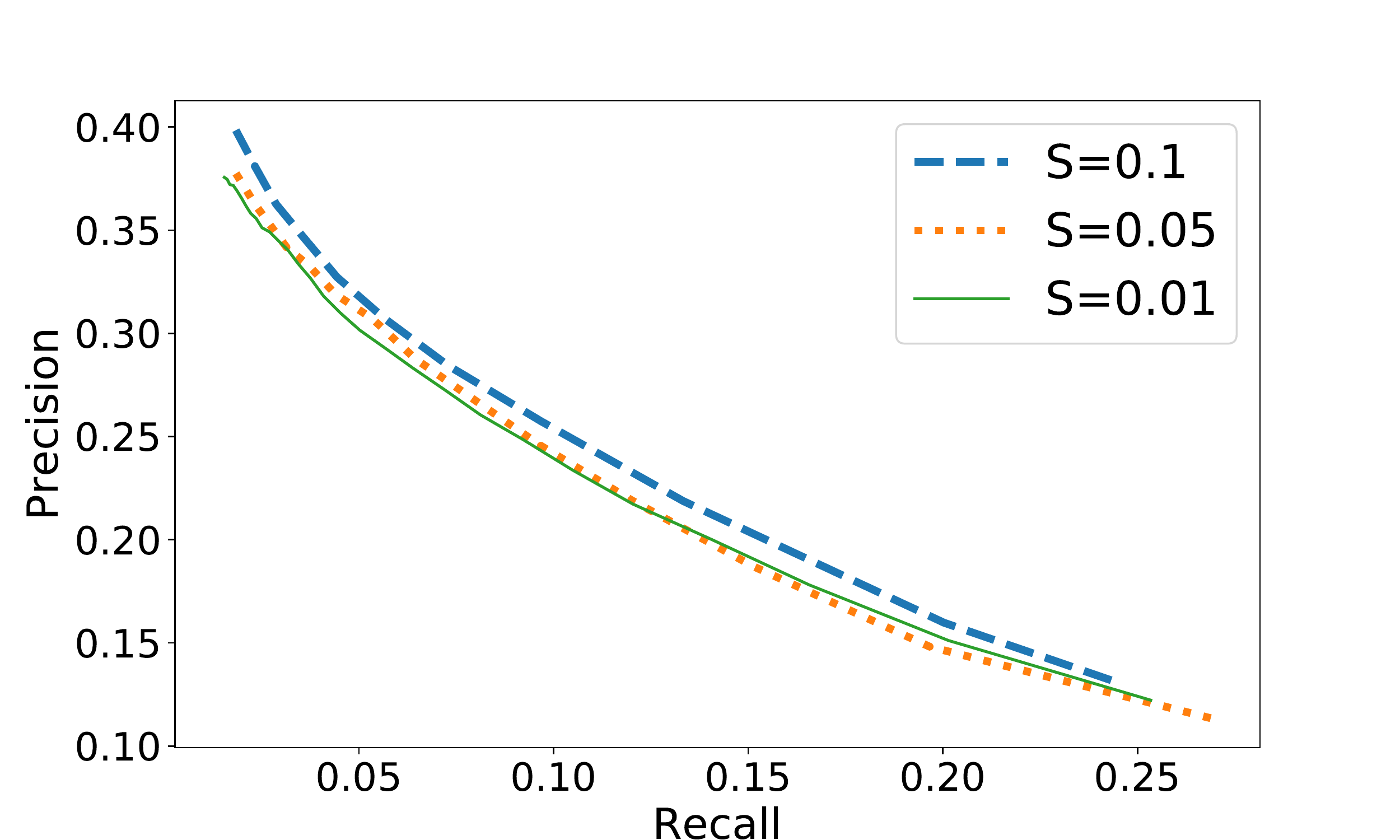}\vspace{5pt}
		\end{minipage}
	}\hspace{-5pt}
	\subfigure[F1]{\label{fig:S_F1}
		\begin{minipage}[l]{0.49\columnwidth}
			\centering
			\includegraphics[width=1.1\textwidth]{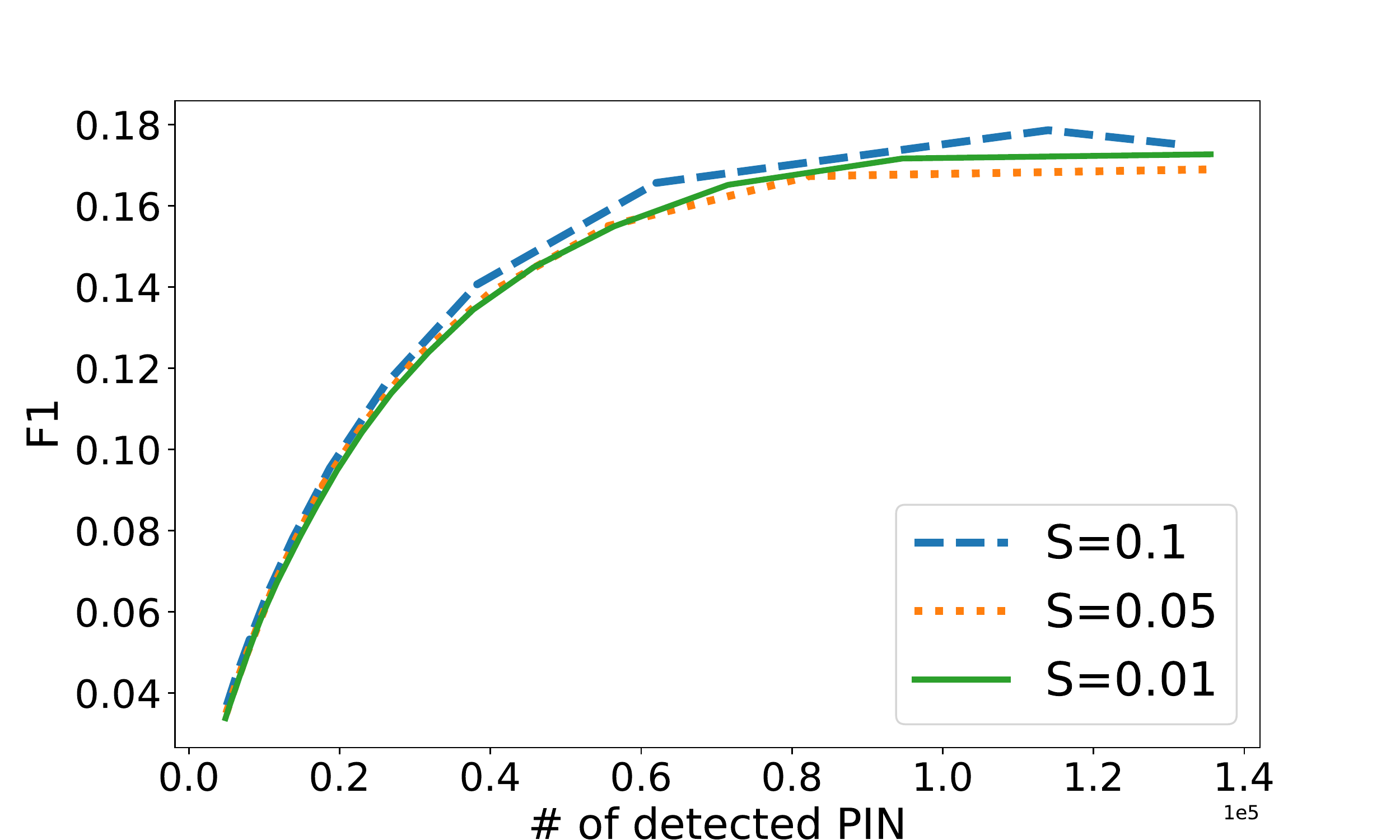}\vspace{5pt}
		\end{minipage}
	}\hspace{-5pt}
	\subfigure[Recall]{\label{fig:S_recall}
		\begin{minipage}[l]{0.49\columnwidth}
			\centering
			\includegraphics[width=1.1\textwidth]{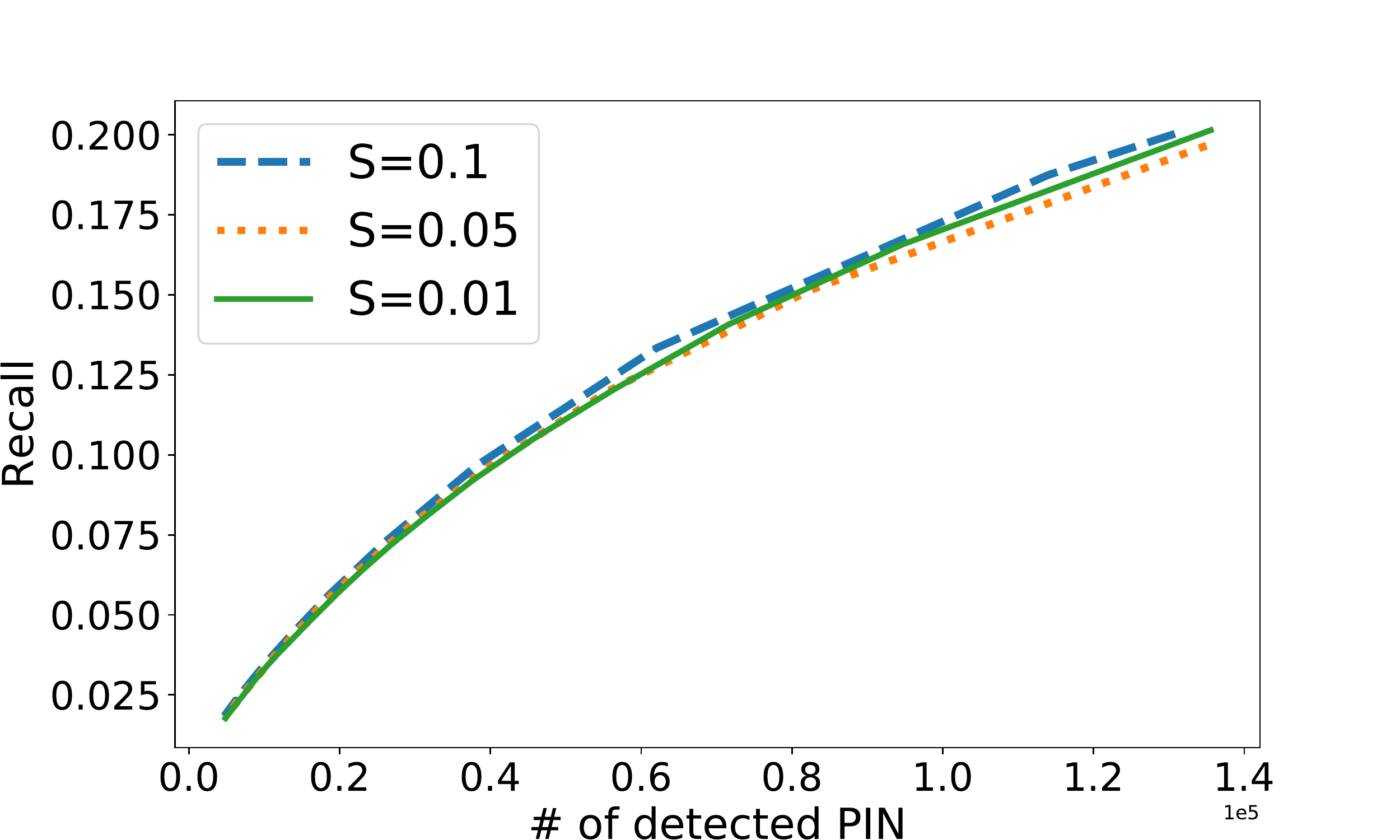}\vspace{5pt}
		\end{minipage}
	}\hspace{-5pt}
	\subfigure[Precision]{\label{fig:S_precision}
		\begin{minipage}[l]{0.48\columnwidth}
			\centering
			\includegraphics[width=1.1\textwidth]{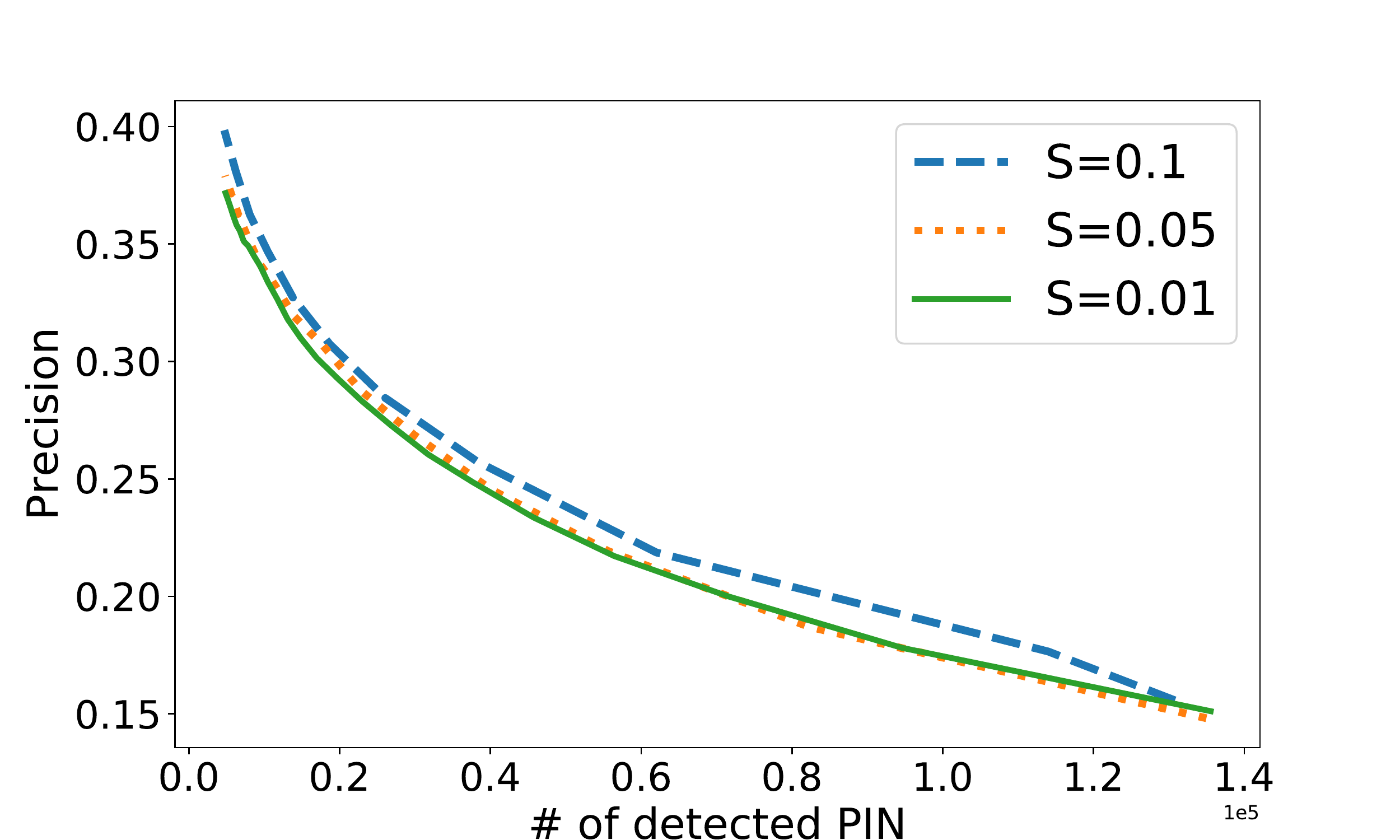}\vspace{5pt}
		\end{minipage}
	}
	%	\vspace{-5pt}
	\caption{Performance Analysis under different $\mathit{S}$ when fixing $\mathit{S}\times\mathit{N}=1$.}\label{fig:parameter_analysis_S}
	%	\vspace{-5pt}
\end{figure*}
%------------------------------------------
%------------------------------------------
\begin{figure*}[h]
	%	\vspace{-10pt}
	\centering
	\subfigure[Precision-Recall curve]{ \label{fig:T_PR_curve}
		\begin{minipage}[l]{0.49\columnwidth}
			\centering
			\includegraphics[width=1.1\textwidth]{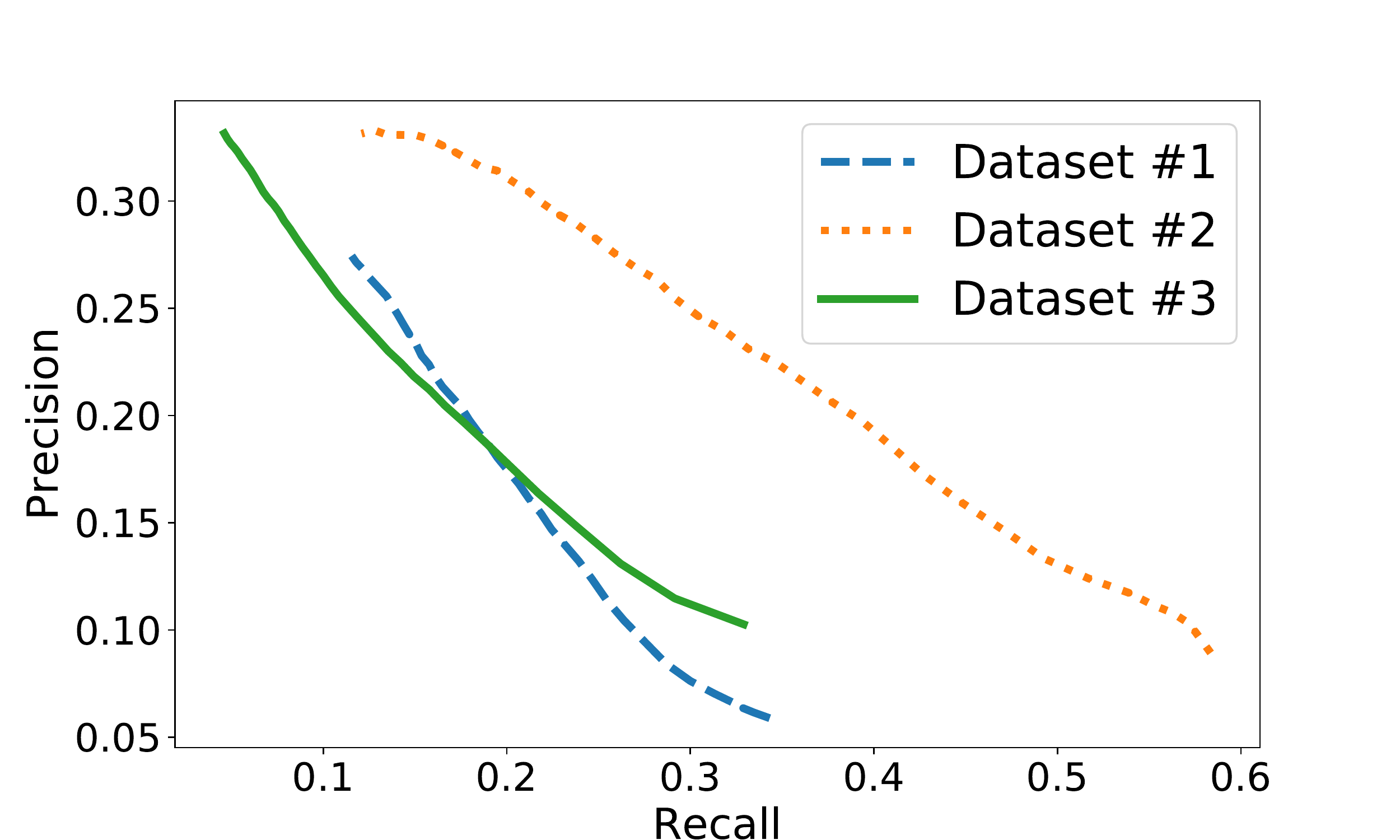}\vspace{5pt}
		\end{minipage}
	}\hspace{-5pt}
	\subfigure[F1]{\label{fig:T_F1}
		\begin{minipage}[l]{0.49\columnwidth}
			\centering
			\includegraphics[width=1.1\textwidth]{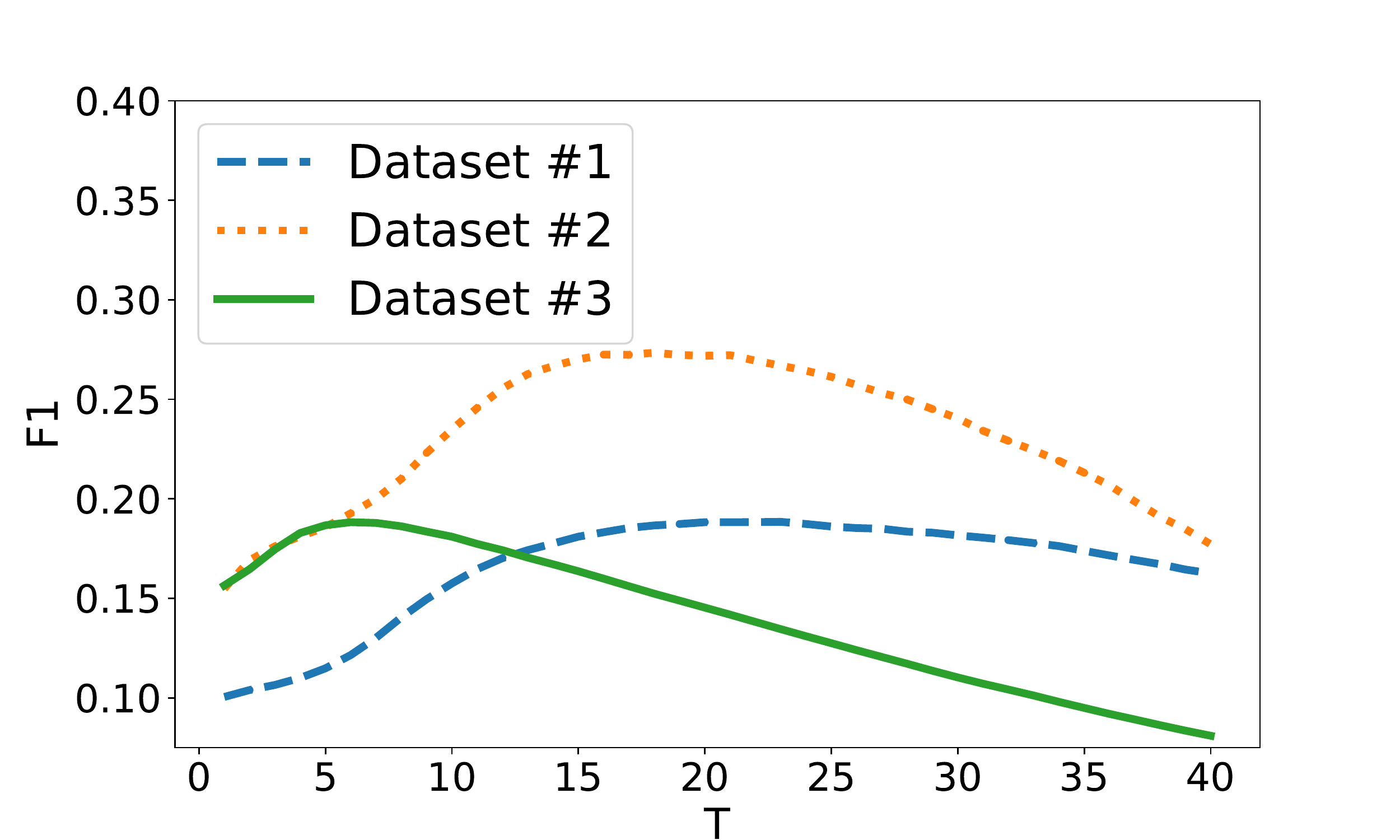}\vspace{5pt}
		\end{minipage}
	}\hspace{-5pt}
	\subfigure[Recall]{\label{fig:T_recall}
		\begin{minipage}[l]{0.49\columnwidth}
			\centering
			\includegraphics[width=1.1\textwidth]{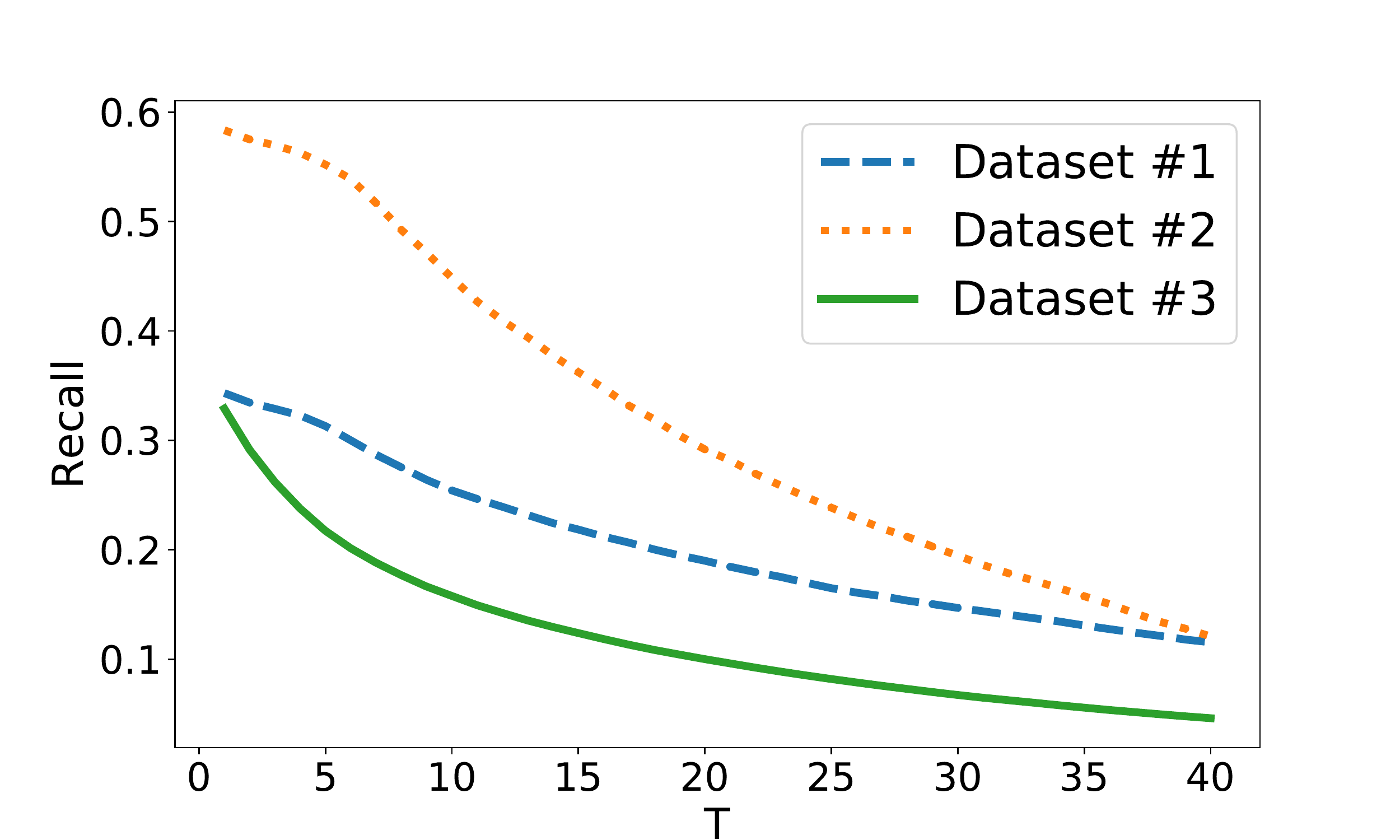}\vspace{5pt}
		\end{minipage}
	}\hspace{-5pt}
	\subfigure[Precision]{\label{fig:T_precision}
		\begin{minipage}[l]{0.48\columnwidth}
			\centering
			\includegraphics[width=1.1\textwidth]{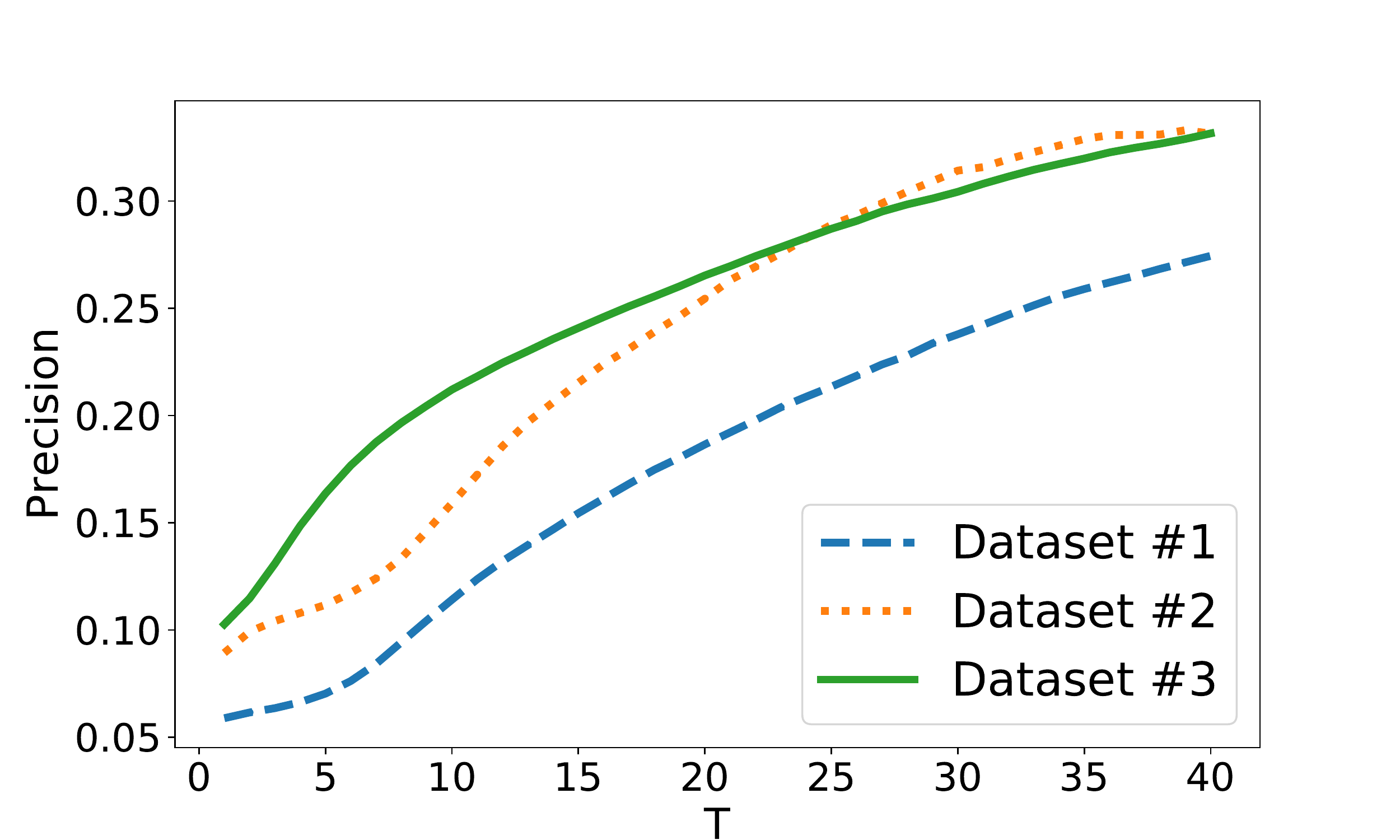}\vspace{5pt}
		\end{minipage}
	}
	%	\vspace{-5pt}
	\caption{Performance Analysis under different $\mathit{T}$ when fixing $\mathit{S}=0.1$ and $\mathit{N}=80$.}\label{fig:parameter_analysis_T}
	%	\vspace{-10pt}
\end{figure*}
%------------------------------------------
%\vspace{-5pt}
\subsection{Impacts of various parameters}\label{sec:parameter}
%\vspace{-2pt}
In this section, we evaluate the impacts of parameters shown in Table~\ref{tab:parameters}. 
% , and the F1 and running time are two main aspects for evaluation. We tested the impact of $\mathit{N}$, and then the sample ratio $\mathit{S}$ was analyzed. At last, voting threshold $\mathit{T}$ was evaluated as well. 
We conducte experiments, where the sampling method is fixed as \textit{RES}, in all three datasets, but only show the results on the No.3 dataset for the reason of the pages limitation when analyzing $\mathit{N}$ and $\mathit{S}$. It should be emphasized that the impacts are consistent across the three datasets in our experiments.

\subsubsection{Impact of $\mathit{N}$}
%\noindent \textbf{(\RNum{1})} \textbf{Impact of $\mathit{N}$}
In order to evaluate the impact of the parameter $\mathit{N}$, we fixed $\mathit{S}$ as $0.1$, and $\mathit{N}$ changed within \{10, 20, 40, 80\}. The Precision-Recall curve, F1, Precision and Recall are displayed in Figure~\ref{fig:parameter_analysis_N}. Here we have to explain why the number of detected nodes is used in the x-axis of Figure~\ref{fig:N_F1}-~\ref{fig:N_precision}. 
Obviously, we doesn't mention the control of the last parameter $\mathit{T}$ consistent in the comparative experiments. In fact, when the parameter $\mathit{N}$ is not the same, the consistent $\mathit{T}$ in not reasonable instead, because in this case there is a huge difference in the total number of votes behind the same $\mathit{T}$. Thus for the sake of fairness, we compare performance when {\our} detects the equivalent fraud nodes under different $\mathit{N}$. 

The results show that {\our} achieves better performance with a rise of $\mathit{N}$. In fact, this elevation of performance comes directly from the nature of the bagging method. However, we should note that the improvement in performance is not significant, especially with the increase of $\mathit{N}$. When comparing $\mathit{N}=40$ and $\mathit{N}=80$, we can find the improvement has become negligible. At the same time, the rise of the cost of equipments is enormous which is not a fair trade-off. In addition, we can also discover that {\our} is very stable when $\mathit{N}\in \{10, 20, 40, 80\}$ which means the repetition rate $\mathit{R}$ is between $1$ and $8$ times. The stability actually makes {\our} has loose requirements for the computational environment: even if there are not enough parallel computing cores, {\our} can still achieve relatively stable and acceptable performance.

\subsubsection{Impact of $\mathit{S}$}
%\noindent \textbf{(\RNum{2})} \textbf{Impact of $\mathit{S}$} 
To evaluate the impact of $\mathit{S}$, we fixed $\mathit{S}\times\mathit{N}=1$, instead of setting $\mathit{N}$ as a constant value. The reason why we choose such an experimental setting is that the same repetition rate $\mathit{R}$ is fairer for points and edges in the bipartite graphs. Besides, $\mathit{R}$ itself is determined by the $\mathit{S}$ and $\mathit{N}$, so after the analysis of $\mathit{S}$ and $\mathit{N}$, no additional analysis is needed for the dependent variable $\mathit{R}$. 
The Figure~\ref{fig:parameter_analysis_S} shows the Precision-Recall curve, F1, Precision and Recall in the experiments with $\mathit{S}\in\{0.01,0.05,0.1\}$. From Figure~\ref{fig:S_PR_curve}-~\ref{fig:S_precision}, we have two major findings. On the one hand, we can see that the rise of $\mathit{S}$ can bring a certain improvement in performance. We should remind that $\mathit{S}$ will determine the size of sampled subgraphs, and the graphs in the real-world have a very large scale normally. A sampled subgraph with a relatively large scale still challenges the storage structure and the single computing core which is against the original intention of {\our}. Thus we don't pay much attention to the performance improvement with the rise of $\mathit{S}$. On the other hand, stability is also shown in this set of experiments. When $\mathit{S}=0.01$, the performance shown in Figure~\ref{fig:parameter_analysis_S} is still close to the one of $\mathit{S}=0.1$. It means that when facing a large-scale graph structure, the stability of {\our} allows you to sample the graph to a much smaller size without losing a lot of performance. Of course, when sampling large-scale graphs to ones of smaller size, $\mathit{N}$ will increase to keep $\mathit{R}$ constant. But we think this trade-off can be done according to task requirements and equipment conditions. If you are more concerned about time consumption with enough parallel computing cores or the original graph is too large to deal with, you can set a smaller $\mathit{S}$. Otherwise, if the performance is more critical or the original graph itself is not too large to handle, a relatively large $\mathit{S}$ is a good choice. In fact, parameters $\mathit{S}$ and $\mathit{N}$ let {\our} very flexible enough to adapt to a variety of scenarios rather than make {\our} complicated to be manipulated.

%\vspace{-8pt}
\subsubsection{Impact of $\mathit{T}$}
%\noindent \textbf{(\RNum{3})} \textbf{Impact of $\mathit{T}$} \ 
% When evaluating the threshold $\mathit{T}$, $\mathit{S}$ and $\mathit{N}$ are set constant.
The results shown in Figure~\ref{fig:parameter_analysis_T} come from the set of experiments with $\mathit{S}=0.1$, $\mathit{N}=80$ and $\mathit{T}\in\{1,2,\ldots\,39,40\}$. Obviously, the experiment results show that Precision would go up and Recall would drop with the rise of $\mathit{T}$. The phenomenon is easy to understand because the fraud nodes with more votes are equivalent to having a higher risk in multiple sampled graphs. Meanwhile, the number of detected nodes will decrease with the increase of the threshold $\mathit{T}$ which necessarily leads to the fall of Recall. We can find the curves are smooth and monotonous in Figure~\ref{fig:T_recall}-~\ref{fig:T_precision} which is a nice property that can be exploited. Based on the curves, we can determine $\mathit{T}$ in response to the task requirements: do we prefer to reduce the detecting error rate or to try to find the fraud nodes as many as possible. In this case, we have a definite direction when we need to tune the parameter $\mathit{T}$.

%-----------------------------------------------
%\vspace{-10pt}
\section{Conclusion}\label{sec:conclusion}
In this paper, we propose an ensemble approach {\our} to solve the promotional campaigns fraud detection problem in large graphs. We first formulate the optimization problem for promotional campaigns fraud detection according to our business scenarios. {\our} scales up fraud detection through the ensemble framework, and we analyze the sampling methods in bipartite graphs as well. A heuristic method {\ouralgorithm}, which is one critical part of {\our}, is proposed to detect fraud subgraphs. The {\ouralgorithm} can automatically search the number of fraud subgraphs, which is a hyper-parameter for previous methods, and speed up the search process through an efficient truncation. Extensive experiments conducted on three real-world datasets demonstrate that {\our} is effective, scalable, and stable. After successful experiments on real data, {\our} has been deployed in the risk control department of JD.com for further tests. 
\balance
\bibliographystyle{plain}
\bibliography{reference}

\end{document}